\documentclass[runningheads]{llncs}

\usepackage{eccv}

\usepackage{eccvabbrv}

\usepackage{graphicx}
\usepackage{booktabs}

\usepackage[accsupp]{axessibility}  % Improves PDF readability for those with disabilities.

\usepackage[pagebackref,breaklinks,colorlinks,citecolor=eccvblue]{hyperref}
\usepackage{orcidlink}

\usepackage{tabularx}
\usepackage{multirow}
\usepackage{colortbl}

\usepackage{amssymb}% http://ctan.org/pkg/amssymb
\usepackage{pifont}% http://ctan.org/pkg/pifont
\definecolor{mycyan}{rgb}{0,0.76,0.75}
\definecolor{LightGrey}{rgb}{0.92,0.92,0.92}

\newcommand{\tabbar}[2]{\rowcolor{gray!15} \multicolumn{#1}{l}{\textit{#2}}}

\begin{document}

\title{Open-Vocabulary 3D Semantic Segmentation with Text-to-Image Diffusion Models} 

\titlerunning{Open Vocabulary 3D  Segmentation with Text-to-Image Diffusion Models}

\author{Xiaoyu Zhu\inst{1\dag} \and
Hao Zhou\inst{2} \and
Pengfei Xing\inst{2} \and
Long Zhao\inst{2} \and
Hao Xu\inst{3} \and
Junwei Liang\inst{4} \and
Alexander Hauptmann\inst{1} \and
Ting Liu\inst{2} \and
Andrew Gallagher\inst{2}}

\authorrunning{X. Zhu et al.}

\institute{Carnegie Mellon University \and
Google DeepMind \and
Google \and
HKUST}

% \maketitle

\maketitle
\begin{center}
% \vspace{-1mm}
  \includegraphics[width=1.0\linewidth]{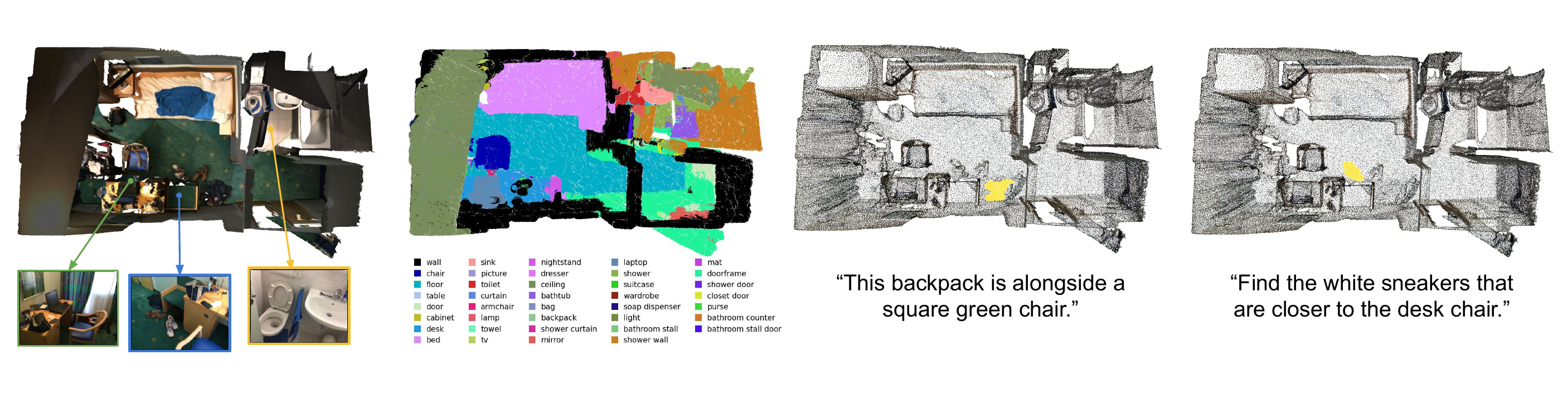}
% \vspace{-2mm}
\captionof{figure}{\textbf{Illustration of open-vocabulary 3D semantic scene understanding}. We propose \emph{Diff2Scene}, a 3D model that performs open-vocabulary semantic segmentation and visual grounding tasks given novel text prompts, without relying on any annotated 3D data. By leveraging discriminative-based and generative-based 2D foundation models, \emph{Diff2Scene} can handle a wide variety of novel text queries for both common and rare classes, like ``desk'' and ``soap dispenser''. It can also handle compositional queries, such as ``find the white sneakers that are closer to the desk chair.''}
\label{fig:fig1}
\end{center}

\begin{abstract}
In this paper, we investigate the use of diffusion models which are pre-trained on large-scale image-caption pairs for open-vocabulary 3D semantic understanding. We propose a novel method, namely Diff2Scene, which leverages frozen representations from text-image generative models, along with salient-aware and geometric-aware masks, for open-vocabulary 3D semantic segmentation and visual grounding tasks. Diff2Scene gets rid of any labeled 3D data and effectively identifies objects, appearances, materials, locations and their compositions in 3D scenes. We show that it outperforms competitive baselines and achieves significant improvements over state-of-the-art methods. In particular, Diff2Scene improves the state-of-the-art method on ScanNet200 by $12\%$.
\end{abstract}
\let\thefootnote\relax\footnotetext{\dag This work was partially done while the author was a student researcher at Google.}    
\section{Introduction}
\label{sec:intro}

3D semantic scene understanding, with the task of assigning semantics to every 3D point, plays a fundamental role in many computer vision applications, such as robotics \cite{xu2020learning}, autonomous driving \cite{huang2020multi}, human-computer interaction \cite{fan2022vision}, and augmented reality \cite{han2020live}.
Traditional studies in this field usually target solving this problem in a closed-set fashion \cite{mask3d,choy20194d}, resulting in models that can only be used to make predictions within the predefined label space.

Recent progress in computer vision have witnessed the emerging interests in solving semantic understanding tasks in open-vocabulary settings \cite{openscene,openmask,roh2022languagerefer,huang2022multi,zhou2022maskclip}.
In contrast to closed-set setting, models targeting open-vocabulary tasks must make predictions for any semantics described in text, including object category and fine-grained attributes (\emph{e.g.}, shape, color, material, property) as well as their complicated compositions. 
However, this is a challenging task due to the wide diversity and complexity of possible queries.
Motivated by the advance of aligning text and image embeddings with large-scale foundation models \cite{clip,Li2021AlignBF,align,alayrac2022flamingo}, existing methods mitigate this challenge by lifting the image features from foundation models such as CLIP \cite{clip} or their descendants \cite{openseg, li2022languagedriven} to 3D.
These lifted feature representations for 3D points can then be used to query with open-vocabulary descriptions, achieving semantic understanding in 3D.
Despite these achievements, contrastively trained CLIP-based models exhibit limitations in handling fine-grained classes \cite{radford2021learning} and novel compositional text queries \cite{ma2023crepe}, restricting their performance in open-vocabulary 3D semantic understanding.

The recently developed text-to-image diffusion models have shown outstanding abilities for image generation even with challenging text prompts \cite{rombach2021highresolution,schuhmann2022laion,liu2022compositional}, such as combinational descriptions with multiple attributes (\emph{e.g., A bucket bag made of blue suede with intricate golden paisley patterns.}) 
The internal visual representation of these models, entangled with text embedding through cross-attention, have proven correlate well with semantic concepts described by language \cite{pmlr-v202-mittal23a,Yang_2023_ICCV,kwon2023diffusion,michele2021generative}. On the other hand, CLIP-based foundation models have been shown to struggle with compositionality \cite{ma2023crepe}.
Moreover, compared with the CLIP model which is optimized for global representation, diffusion models have proven to be superior at local representation \cite{tang2023dift}, which is a key for dense prediction tasks.
Specifically, ODISE \cite{odise} applied the internal representations of Stable Diffusion \cite{rombach2021highresolution} to open-vocabulary 2D semantic understanding tasks and achieved promising results.

One of the key challenges in 3D perception is the severe scarcity of point clouds and their dense labels. 
Several existing methods have been proposed to solve the lack of data issue in a zero-shot fashion by leveraging the CLIP model pre-trained on large-scale text-image data \cite{openscene,takmaz2023openmask3d,conceptfusion}.
The prior art \cite{openscene} extracts dense CLIP features from 2D images and distill the knowledge of their lifted 3D counterpart into a 3D mask predictor.
However, CLIP features, as discussed above, struggle to handle fine-grained classes \cite{radford2021learning} and show worse localization capability compared with diffusion features.
We leverage diffusion model as feature backbone along with a mask-based segmentation head (\emph{e.g.}, Mask2Fromer~\cite{cheng2021mask2former})
for its intrinsic nature that decouples mask and its semantic representations.
This is intuitively suitable for leveraging semantically-rich embeddings from 2D foundation models and further learning geometrically-accurate masks from the 3D branch. 
However, performing multi-modal distillation with mask-based segmentation head is a non-trivial task.
The frozen features extracted from the decoder of the U-Net in the diffusion model are trained with generative objectives, and cannot be directly used for the perception task.
Therefore, directly distilling knowledge from these features as normally done in prior art \cite{openscene, 3dovs, liu20213d} is infeasible.
Another intuitive way is to leverage a supervised 3D mask proposal network and pool the feature representations from 2D CLIP features for each mask \cite{takmaz2023openmask3d}.
However, the training of 3D mask proposal network requires labeled 3D data, which may not be feasible in practice.

\begin{figure*}[t]
% \hspace*{-0.5cm}    
\includegraphics[width=1.05\linewidth]{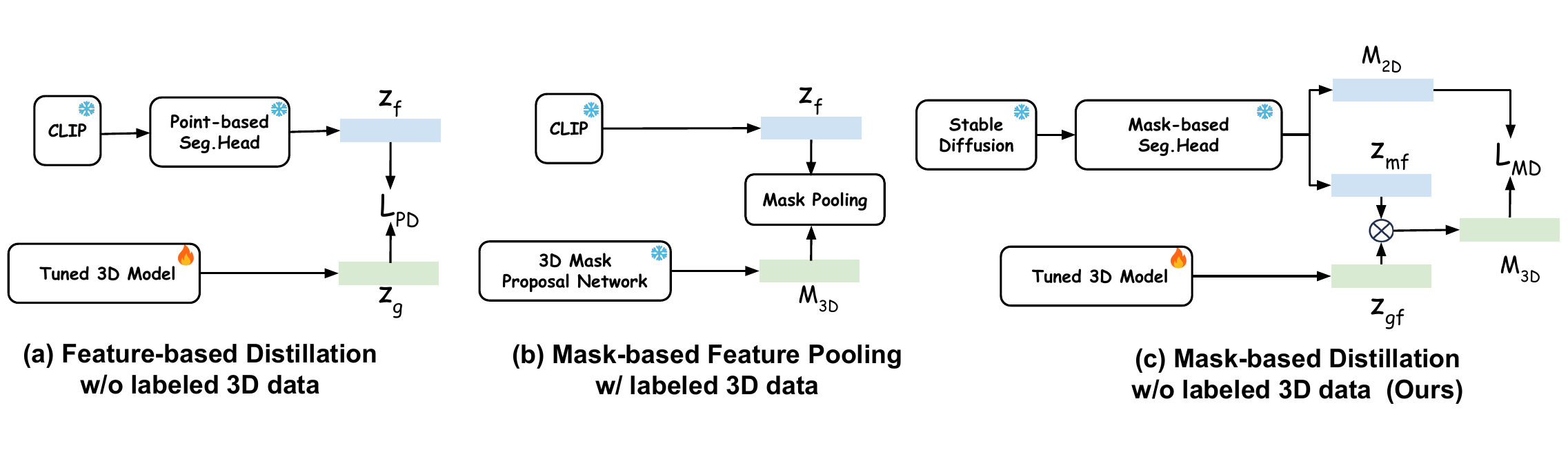}
\caption{
\textbf{Illustration of open-vocabulary 3D perception methods.} $L_{PD}$ and $L_{MD}$ denote point-based distillation loss and mask-based distillation loss. $M_{3D}$ denote a set of predicted 3D masks; $M_{2D}$ and $Z_{mf}$ denote a set of predicted 2D masks and their semantic embeddings; $Z_{gf}$ denote the high-resolution 3D feature map. (a) Directly minimizing the per-point feature distance between the CLIP-based model and the tuned 3D model \cite{openscene}. (b) Directly using a 3D mask proposal network trained on labeled 3D data to produce class-agnostic masks, and then pool corresponding representations from the CLIP feature map \cite{takmaz2023openmask3d}.  (c) The proposed mask distillation approach, namely \emph{Diff2Scene}, that uses Stable Diffusion and performs mask-based distillation. \emph{Diff2Scene} leverages the semantically-rich mask embeddings from 2D foundation models and geometrically accurate masks from the tuned 3D model, and thus achieves superior performance compared to previous methods. 
\textcolor{red}{}
} 
\label{fig:overview}
\end{figure*}

To mitigate these issues, we propose a novel mask distillation method tailored to distill knowledge from the Mask2Former style 2D branch \cite{cheng2021mask2former, odise} to the 3D branch, which is shown in Fig.~\ref{fig:overview}.
Specifically, we design our 3D branch to take a 3D point cloud as input and to predict their 3D features.
The semantically meaningful mask embeddings produced from our 2D branch are used as linear classifiers to assign class probability to these 3D features.
Their corresponding 2D masks are lifted to 3D based on pixel-point correspondence and used to force the consistency learning of the 2D and 3D branch.

We evaluate \emph{Diff2Scene} quantitatively on ScanNet \cite{dai2017scannet}, ScanNet200 \cite{rozenberszki2022language}, Matterport 3D \cite{Matterport3D} and Replica \cite{straub2019replica} for open-vocabulary 3D semantic segmentation and qualitatively on Nr3D \cite{achlioptas2020referit_3d} for visual grounding tasks.
Our experimental results show that \emph{Diff2Scene} outperforms state-of-the-art models \cite{openscene} on all the four semantic segmentation datasets and achieves promising results on visual grounding tasks.
In summary, we make the following contributions:
\begin{itemize}
    \item To the best of our knowledge, we are the first to leverage text-image diffusion to perform open-vocabulary 3D semantic segmentation. 
  \item We propose a novel mask distillation method to train a 3D mask prediction model by distilling knowledge from the Mask2Former style 2D segmentation model.
  \item The proposed method achieves state-of-the-art performance on several open-vocabulary 3D semantic segmentation and visual grounding benchmarks. 
\end{itemize}

\section{Related Work}
\label{sec:related}

\textbf{Closed-vocabulary 3D semantic segmentation.}
In 3D semantic segmentation, a semantic category is assigned for each 3D point.
It has been long studied \cite{Anand2011,Iro2016,Atzmon2018,3dsemseg_ICCVW17,Engelmann_2019,Graham2018,hu2021vmnet,Hua2018,Koppula2011,Landrieu2018,Li2018,Lu2012,segcloud,wang2015,pointnet,pointnet++,zhu2021weakly} due to its importance in computer vision and robotics applications.
One challenge of this task is that 3D point clouds are not in a regular structured format; network architectures that work well for 2D tasks cannot handle 3D point clouds effectively.
As a result, most of the early studies focus on designing effective and efficient network architectures that are suitable for 3D point clouds \cite{choy20194d,3dsemseg_ICCVW17,Engelmann_2019,Graham2018,hu2021vmnet,Hua2018,Landrieu2018,pointnet,pointnet++,kpconv}.
This line of work achieved great success and significantly improves the results of 3D semantic segmentation. %
Another challenge is the lack of large scale data with ground truth annotations.
Due to the intensive labeling effort and high cost of data annotation \cite{rozenberszki2022language}, the available datasets for 3D semantic segmentation are usually small in scale.
In the absence of large scale data, early studies usually target solving this problem in a closed-vocabulary setting, where the trained model can only predict categories that appear during training.
To mitigate the scale limitation of existing datasets, a handful of works \cite{Cheraghian2019MitigatingTH,Cheraghian2020,cheraghian2019L,cheraghian2022ZSL,zhang2021pointclip,michele2021generative,liu2022languagelevel} have applied zero-shot learning in 3D scene understanding tasks.
\cite{Cheraghian2019MitigatingTH,Cheraghian2020,cheraghian2019L,cheraghian2022ZSL,zhang2021pointclip} focused on 3D point classification task and \cite{michele2021generative,liu2022languagelevel} tried to address the 3D semantic segmentation problem. 
However, these zero-short methods still require ground truth annotations for a certain amount of 3D point clouds. 

\noindent\textbf{Open-vocabulary 3D segmentation.}
The recent progress of large-scale vision and language representation learning \cite{clip,Li2021AlignBF,align,alayrac2022flamingo,he2024unim,zhang2023clipfo3dlearningfreeopenworld} has advanced the study of semantic and instance segmentation in an open-vocabulary setting.
\cite{li2022languagedriven,openseg,ov-seg} first explored open-vocabulary 2D semantic segmentation.
They proposed aligning per-pixel features \cite{li2022languagedriven} or features from mask regions \cite{openseg,ov-seg} with the corresponding text embedding.
Following these works, \cite{chen2022openvocabulary,ding2022language,gadre2022cow,huang23vlmaps,conceptfusion,Mazur:etal:ICRA2023,openscene,shafiullah2022clipfields,shah2022lmnav} focus on 3D semantic segmentation in an open-vocabulary setting.
Among them, \cite{chen2022openvocabulary,gadre2022cow,huang23vlmaps,conceptfusion,Mazur:etal:ICRA2023,shafiullah2022clipfields,saharia2022photorealistic} project 3D points to 2D images and solve the 3D problem in the 2D space, instead of targeting the 3D open-vocabulary semantic segmentation directly.
As \cite{ding2022language} pointed out, the projection from 3D to 2D has information loss and the solution is suboptimal.

To make better use of information from the 3D point cloud, \cite{ding2022language}, \cite{lowis3d} and \cite{openscene} proposed to directly applying semantic segmentation on the 3D point cloud.
\cite{ding2022language} and its extension \cite{lowis3d} proposed associating captions generated for 2D images to  corresponding 3D point clouds to build the pseudo-ground truth captions for 3D point clouds.
A neural network is trained to associate the 3D point cloud with these pseudo labels through contrastive loss.
Similar to the zero-shot setting, \cite{ding2022language} evaluated their model in a leave-one-out fashion, which still requires annotations for 3D point cloud.
Inspired by the strong open-vocabulary ability of large-scale vision and language models, Peng \etal \cite{openscene} proposed distilling  knowledge to a 3D point cloud model.
They trained 3D semantic segmentation model by only distilling the knowledge from a CLIP-style~\cite{clip} 2D open-vocabulary semantic segmentation model \cite{li2022languagedriven,openseg}.
They demonstrated that without training with any ground truth labels, the model can achieve great performance on many open-vocabulary tasks.
However, we observe that \cite{openscene} is strongly limited by the 2D open-vocabulary semantic segmentation models used as the teacher.
Its performance on rare classes that are not used in training these models are not satisfactory.
Our method follows this idea by distilling the knowledge of 2D open-vocabulary semantic segmentation model to a 3D model.
In constrast to the approach in \cite{openscene}, we use a diffusion-based 2D open vocabulary semantic segmentation model \cite{odise} as the teacher model.

\noindent\textbf{Diffusion models for scene understanding.}
The last few years have witnessed the success of diffusion models in image generation \cite{rombach2021highresolution,imagen}.
Recent studies also observed the diffusion models are strong representation learners \cite{pmlr-v202-mittal23a,Yang_2023_ICCV,kwon2023diffusion,michele2021generative}.
As a result, researchers have applied it to many understanding tasks such as image classification \cite{Li_2023_ICCVa}, object detection \cite{Chen_2023_ICCV}, image semantic segmentation \cite{baranchuk2022labelefficient,odise,Ji_2023_ICCV}, instance segmentation \cite{Li_2023_ICCV}, human pose estimation \cite{Holmquist_2023_ICCV,Shan_2023_ICCV,Feng_2023_ICCV}, action segmentation \cite{Liu_2023_ICCV}, camera pose estimation \cite{Wang_2023_ICCV}, to name a few, and achieved great success.
Especially, \cite{odise} and \cite{Li_2023_ICCV} showed that Stable Diffusion \cite{rombach2021highresolution}, whose internal representation being well correlated with text embedding, has strong open-vocabulary abilities for understanding tasks.
Inspired by this, we are the first to apply text-to-image diffusion models to open vocabulary 3D semantic segmentation task.
\section{Diff2Scene}

\begin{figure*}[t]
\centering
\includegraphics[width=\linewidth]{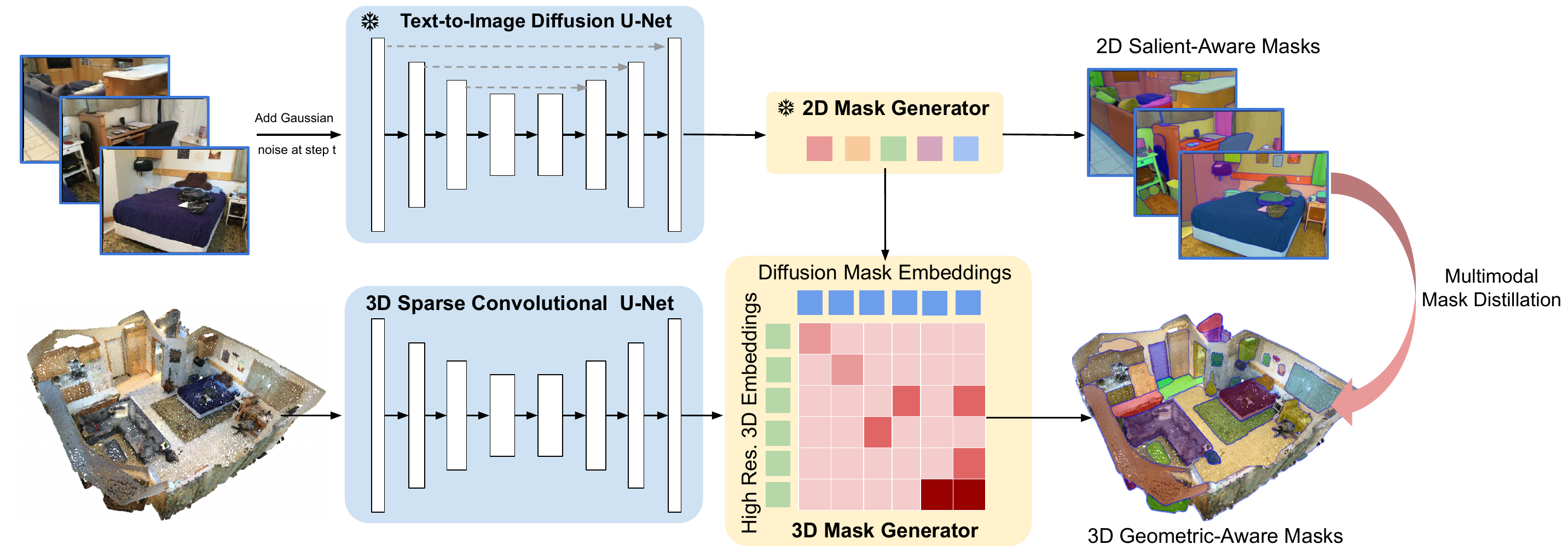}%
\caption{
\textbf{Overview of our method.} We propose \emph{Diff2Scene}, an open-vocabulary 3D semantic understanding model. \emph{Diff2Scene} contains two branches. The 2D branch is designed to be a diffusion-based 2D semantic segmentation model. It accepts a 2D image as input and predicts a set of 2D probabilistic masks with corresponding semantically-rich mask embeddings. The 3D branch utilizes the point cloud and 2D mask embeddings as input. The 2D mask embeddings are used as ``semantic queries'' to generate corresponding 3D probabilistic masks. The model learns salient patterns from the RGB images and geometric information from the point clouds.
\textcolor{red}{}
} 
\label{fig:method}
\end{figure*}

We introduce \emph{Diff2Scene}, an open-vocabulary 3D semantic understanding method.
Similar to \cite{openscene}, our proposed model operates in a zero-shot fashion, where no ground truth 3D annotations are needed during training.
\subsection{Overview}
An overview of \emph{Diff2Scene} is shown in Fig.~\ref{fig:method}.
It takes posed RGB images and the reconstructed 3D point cloud as model inputs. The model predicts the semantic label for each 3D point.
\emph{Diff2Scene} has two branches. The 2D branch is designed to be an open-vocabulary 2D semantic segmentation model.
It leverages text-to-image generative model \cite{rombach2021highresolution} which is pre-trained on massive text-image pairs.
The model takes a 2D image as input to predict a set of 2D probabilistic masks with their corresponding 2D mask embeddings.
Thanks to the generative pre-training process with large-scale text-image pairs, the 2D mask embeddings are semantically rich.
The model leverages the salient patterns in RGB images to produce the 2D salient masks.
The 3D branch takes the point cloud and the 2D mask embeddings as inputs.
The 2D mask embeddings are used as linear classifiers to assign class probabilities to each of 3D features output from the 3D branch, resulting in a 3D probabilistic mask termed as geometric masks.
To predict the per-point semantic class,  the model first computes the per-mask category logits for both salient and geometric masks.
Then we ensemble the per-mask logits for those two types of masks. In the way, the model can learn salient patterns from the RGB images and geometric information from the point clouds.

\subsection{2D Semantic Understanding Model}
One challenge of 3D semantic understanding is the severe scarcity of 3D point clouds with groundtruth labels.
To tackle the challenge brought by limited training data, vision-language foundation models have been used to transfer semantically-rich 2D features into the 3D space \cite{takmaz2023openmask3d, openscene, conceptfusion}. \cite{takmaz2023openmask3d} used on a model trained on labeled 3D data to produce class-agnostic masks, and then pooled the corresponding 2D representations as the mask embeddings. On the other hand, \cite{openscene} proposed to leverage a pre-trained 2D semantic segmentation model as feature extractor to perform open-vocabulary 3D segmentation, and no ground truth 3D annotations are needed during training. In this work, we follow the setting in \cite{openscene} to reduce the 3D annotation efforts.  

The 2D segmentation model consists of an image backbone $\phi$ which is a foundation model pretrained on large-scale text-image pairs; and a segmentation head $\sigma$ to predict the semantic embedding.  There are multiple design options for the 2D backbone $\phi$ and segmentation head $\sigma$.  (1) The 2D backbone could either be \textbf{contrastively pretrained} or \textbf{generatively pre-trianed}. 
The popular frameworks for contrastive representation learning include CLIP \cite{clip} and ALIGN \cite{jia2021scaling}. On the other hand, a few works \cite{odise,zhao2023unleashing,wang2023diffusion} have demonstrated promising performance by using generatively pre-trained representations for perception task. The feature representations from Diffusion U-Net blocks are extracted for different downstream tasks.

Once feature representations from text-image foundation models are extracted, a segmentation head is added upon those features to predict the per-point semantic classes. The segmentation problem could be formulated as \textbf{pixel-based classification} or \textbf{mask-based classification}. For pixel-based classification \cite{li2022languagedriven,openseg}, the intermediate output of segmentation head is of shape $H\times W \times C$, where $H$ and $W$ is image height and width, and $C$ is the dimension of feature embedding. For mask-based classification \cite{cheng2021mask2former,cheng2021maskformer, odise}, the segmentation head takes the 2D feature map $\mathbf{F}^{2d}$ and $N$ fixed mask queries $\{q_i\}_{i=1}^N$ as input. The intermediate output is $N$ 2D probabilistic masks $\{\mathcal{B}^{2d}_i\}_{i=1}^N$ and their corresponding mask embeddings $\{f^{2d}_i\}_{i=1}^N$.

In this work, we choose diffusion model as the feature backbone $\phi$, considering its strong localization ability brought by generative pre-training. Besides, we leverage mask-based segmentation head for its intrinsic nature that decouples mask and its semantic representations. This is intuitively suitable for leveraging semantically-rich embeddings from 2D foundation models, and further learn geometrically-accurate masks from the 3D branch.

\subsection{Geometry-Aware 3D Mask Model}

While mask-based segmentation has achieved promising performance in fully-supervised setting \cite{cheng2021mask2former,cheng2021maskformer,mask3d}, it has been rarely explored to transfer the learned mask-level representations into another domain.
On the other hand, the point-based feature representations from 2D foundation model can be naively distilled by minimizing the per-point feature distance. For example, \cite{openscene} proposed to train a 3D model to predicts 3D semantic meaningful features by distilling pixel aligned 2D features.
However, similar methods are not applicable in our proposed method.
First of all, our 2D semantic understanding model uses a mask-based segmentation head which does not provide semantically-rich features in the pixel level.
Secondly, the backbone of our 2D semantic understanding model is a frozen stable diffusion model \cite{rombach2021highresolution} which is designed to generate realistic images with rich details and not tuned for semantic segmentation tasks.
The per-pixel features extracted from it are not feasible to supervise the training of our 3D mask model\textsuperscript{\dag}.\footnote{\dag The 3D mask model trained to distill these features does not converge.}
In the following, we introduce our proposed mask distillation which is tailored to distill knowledge from the mask-based 2D foundation model to the geometry-aware 3D mask model.

The  mask-based 2D foundation model predicts $N$ 2D probabilistic masks $\{\mathcal{B}^{2d}_i\}_{i=1}^N$ and their corresponding mask embeddings $\{f^{2d}_i\}_{i=1}^N$.
Specifically, $\mathcal{B}^{2d}_i$ represents a probabilistic map whose elements represent the probability of the corresponding pixel being foreground.
We first compute the pixel-point correspondence following \cite{openscene}.
Subsequently, a set of 3D probabilistic masks $\{\mathcal B^{3d}_i\}_{i=1}^{N}$ can be generated by lifting the 2D masks $\{\mathcal{B}^{2d}_i\}_{i=1}^N$ to 3D space based on the pixel-point correspondence.
We proposed a novel mask distillation which distills information from both 3D probabilistic masks $\{\mathcal B^{3d}_i\}_{i=1}^{N}$ and the corresponding semantic rich mask embeddings $\{f^{2d}_i\}_{i=1}^N$ generated from the 2D branch.
Specifically, we train a Minkowski network \cite{choy20194d} as the 3D mask prediction model to generate geometry-aware 3D masks.
The 3D point cloud is quantized into voxels by averaging the pixels within each voxel to save memory and reduce computes.
The 3D mask prediction model generates a 3D feature to represent each voxel and this feature is assigned to all points within the voxel.
This produces a full feature map $\mathbf{F}^{3d} \in \mathbb{R}^{M \times D}$ for the point cloud, where $D$ is the dimension of the 3D feature.
The semantic rich 2D mask embeddings $\{f_i^\text{2d}\}_{i=1}^N$ are used as linear classifiers to compute the logits $\mathcal{S}_i \in \mathbb{R}^{M}$ of a 3D feature belonging to the corresponding class:
\begin{equation}
    \mathcal{S}_i = \langle\mathbf{F}^{3d},f^{2d}_i\rangle,
\end{equation}
where $\langle\cdot,\cdot\rangle$ denotes inner product. 
The 3D probabilistic mask $\mathcal{B'}^{3d}_i$ is then generated by applying the sigmoid function on $\mathcal{S}_i$. We propose a multimodal mask distillation loss to train our 3D mask generator:
\begin{equation}
    \mathcal{L} = \sum_{i=1}^{N}1 - \text{cos}(\mathcal{B'}^{3d}_i, \mathcal{B}^{3d}_i).
\end{equation}

The distillation loss aims at forcing the 2D and 3D branch to make consistent predictions. It serves as an implicit distillation objective to make the 3D model learn high-resolution, semantically-rich feature representations.  

\subsection{Open-Vocabulary Inference}
During inference, \emph{Diff2Scene} takes a 3D point cloud and its multiview 2D images as inputs.
The 2D semantic understanding model consumes the 2D images and generates a set of 2D probabilistic masks $\{\mathcal{B}^{2d}_i\}_{i=1}^{N}$ with their corresponding mask embeddings $\{f_i^\text{2d}\}_{i=1}^N$, where $f^{2d}_i \in \mathbb{R}^D$.
The 3D mask model takes the 3D point cloud and the mask embeddings $\{f_i^\text{2d}\}_{i=1}^N$ as inputs to predict the 3D probabilistic mask $\{\mathcal{B'}^{3d}_i\}_{i=1}^{N}$.
To ground a semantic label $c$ to the 3D point cloud, we first apply the same idea from \cite{odise} to compute the geometric mean (denoted as $p_i^c$) of label probabilities from diffusion and discriminative models for each 2D mask $\{\mathcal{B}^{2d}_i\}_{i=1}^{N}$.
Next, the label probabilities $\mathbf{p}^c$ are assigned to 3D points via the following equation:
\begin{equation}
    \mathbf{p}^c = \lambda \sum_{i=1}^N p^c_i*\mathcal{B}^{3d}_i+ (1-\lambda) \sum_{i=1}^N p^c_i*\mathcal{B'}^{3d}_i, \label{eq:probability}
\end{equation}
where $\lambda=0.5$. 
When multiple labels can be assigned to a 3D point, the label with the highest probability from Eq.~\ref{eq:probability} is taken.

\section{Experiment}
We conduct a series of experiments to demonstrate the effectiveness of \emph{Diff2Scene} on a variety of zero-shot 3D scene understanding benchmarks.
% %
We first evaluate the proposed model on zero-shot open-vocabulary semantic segmentation tasks following the evaluation protocol of \cite{openscene}.
% %
We then perform comprehensive ablation studies to validate our designs.
% %
Finally, we qualitatively demonstrate the strong ability of the proposed model for open-vocabulary 3D segmentation and grounding complicated compositional text queries.

\subsection{Datasets}
We use ScanNet \cite{dai2017scannet}, Matterport3D \cite{Matterport3D}, ScanNet200 \cite{rozenberszki2022language} and Replica \cite{straub2019replica} for the open-vocabulary 3D semantic segmentation task. We provide qualitative analysis of the visual grounding task on Nr3D \cite{achlioptas2020referit_3d}.
Except for Replica, point clouds and multi-view images in the training split without ground truth annotations are used for model training.
As Replica does not provide the training data,
we perform training on ScanNet and perform evaluation on Replica, following the setting in \cite{takmaz2023openmask3d}.

\paragraph{ScanNet} is one of the largest 3D semantic segmentation dataset.
It provides 80,554 images from 1201 scans for training and 21,300 images from 312 scans for testing with 20 semantic labels.

\paragraph{Matterport3D}
is a large scale RGB-D dataset containing 10,800 panoramic views from 194,000 RGB-D images of 90 building-scale scenes.
It splits 61 scenes for training, 11 scenes for validation and 18 for testing.
We train our 3D branch using the images in the training splits and report the results on test split.

\paragraph{ScanNet200} has 200 semantic labels with long-tailed classes.
It also provides a grouping of the 200 categories based on the number of labeled surface points in the training set, resulting in 3 subsets: head,
common, and tail.
This enables us to evaluate the performance of our method on the long-tail distribution, making ScanNet200 a natural choice as an evaluation dataset.
We report the mean intersection over union (mIoU) metric on the validation set consisting of 312 scenes following the split in \cite{openmask,openscene,rozenberszki2022language}.

\paragraph{Replica} 
contains 51 categories, and we further split those categories into head and tail sets based on their appearance frequency.
We report the mIoU on the \emph{office0}, \emph{office1}, \emph{office2}, \emph{office3}, \emph{office4}, \emph{room0}, \emph{room1}, and \emph{room2}.

\paragraph{Nr3D} is a 3D visual grounding dataset which contains diverse text prompts. To further evaluate the ability of our model to distinguish between objects in the same class but with different attributes, we perform qualitative evaluation on the visual grounding dataset Nr3D \cite{achlioptas2020referit_3d}.
% %
We perform zero-shot evaluation on the validation set without training on any labeled data for the visual grounding task.

\subsection{Baseline Methods}
We compare \emph{Diff2Scene} with the current state-of-the-art fully-supervised 3D semantic segmentation models including TangentConv \cite{Tat2018}, TextureNet  \cite{huang2019texturenet}, SFSS-MMSI  \cite{choy20194d}, CSC-Pretrain \cite{hou2021exploring}, SupCon \cite{zheng2021weakly}, LGround \cite{rozenberszki2022language} and MinkowskiNet \cite{choy20194d} on the 3D semantic segmentation benchmark.
We also compare our model against OpenScene \cite{openscene} and ConceptFusion \cite{conceptfusion}, the recently proposed open-vocabulary 3D semantic understanding model.
For OpenScene \cite{openscene}, we compare with its OpenSeg \cite{openseg} variant which has the same feature and pre-trained datasets for a fair comparison.
We also compare our model with its three different variants (2D Fusion, 3D Distill, and 2D/3D Ensemble).
Besides, we adapt the state-of-the-art 3D instance segmentation model OpenMask3D \cite{takmaz2023openmask3d} for comparison on the 3D semantic segmentation benchmark. 
%

% \vspace{-5mm}
\begin{table}[!t]
	\centering
	\caption{Comparison to state-of-the-art models. We report mIoU for all benchmarks. Best results in zero-shot, open-vocabulary setting are shown in bold.}
	\resizebox{1.0\textwidth}{!}{\begin{tabular}{l|c|c|cccc|ccc}
	\toprule
	& \textbf{ScanNet} & \textbf{Matterport3D} & \multicolumn{4}{c|}{\textbf{ScanNet200}} & \multicolumn{3}{c}{\textbf{Replica}}  \\
	\midrule
	 & All & All & Head  & Common & Tail  & All & Head & Tail & All\\
	\midrule
	\tabbar{10}{Fully-supervised} \\
    TangentConv \cite{Tat2018}             & 40.9 & - & -  & - & -& -  & - & - & - \\ 
    % \rowcolor{LightGrey}
    TextureNet  \cite{huang2019texturenet} & 54.8 & - & -  & - & -& -  & - & - & - \\ 
    SFSS-MMSI  \cite{choy20194d}         & - & 35.9   & -  & - & -& -  & - & - & - \\ 
    % \rowcolor{LightGrey}
    CSC-Pretrain \cite{hou2021exploring}   & - & - & 45.5  & 17.1 & 7.9& 24.9 & - & - & - \\ 
    % \rowcolor{LightGrey}
    SupCon \cite{zheng2021weakly}           & - & - & 48.6 & 19.2  & 10.3 & 26.0 & - & - & -  \\ 
    % \rowcolor{LightGrey}
    LGround \cite{rozenberszki2022language} & - & - & 48.5 & 18.4     & 10.6     & 27.2  & - & - & - \\ 
    % \rowcolor{LightGrey}
    MinkowskiNet  \cite{choy20194d}         &69.0 & 54.2 & 46.3    & 15.4     & 10.2     & 25.3 & - & - & - \\ 
    \midrule
    \tabbar{10}{Zero-shot, open-vocabulary} \\
	MSeg Voting \cite{MSeg_2020_CVPR}      & 31.0 & 33.4 & -     & -    & -    & -    & -     & -     & -  \\ 
	ConceptFusion \cite{conceptfusion}     & 33.3 & -    & 17.5  & 6.3  & 2.8  & 8.8  & 11.6  & 3.5   & 4.6 \\
    OpenMask3D \cite{takmaz2023openmask3d} & 34.0 & -    & 19.6  & 7.5  & 4.5  & 10.5 & 13.2  & 3.4   & 4.8 \\ 
	OpenScene (2D)  \cite{openscene}       & 41.4 & 32.4 & 21.9  & 10.8 & 5.5 & 12.7 & 33.4  & 11.5  & 14.5 \\
	OpenScene (3D)  \cite{openscene}       & 46.0 & 41.3 & 17.6  & 0.0  & 0.0 & 6.3  & 32.6  & 7.7   & 11.1 \\
	OpenScene (2D/3D)  \cite{openscene}    & 47.5 & 42.6 & 20.0  & 9.7  & 5.1 & 11.6 & 34.2  & 11.9  & 14.9  \\
	\textbf{Diff2Scene (Ours)}   & \textbf{48.6} & \textbf{45.5} & \textbf{25.6}  & \textbf{11.5 }  & \textbf{6.9}  & \textbf{14.2} & \textbf{46.2}  & \textbf{12.9} & \textbf{17.5} \\ 
	\bottomrule
\end{tabular}}
	\label{tab:main_results}
% 	\vspace{-5mm}
\end{table}
\subsection{Implementation Details}
We use posed multi-view RGB images and 3D point clouds for all the datasets.
ODISE \cite{odise}, which consists of a diffusion backbone and mask-based segmentation head, is used as the model in our 2D branch.
It uses a stable diffusion model \cite{rombach2021highresolution} pre-trained on Laion-5B \cite{schuhmann2022laion} as the feature backbone.
The dimensions for diffusion and CLIP features are 256 and 768 respectively.
The number of queries of Mask2Former~\cite{cheng2021mask2former} is 100.
Similar to OpenScene \cite{openscene}, we use MinkowskiNet18A \cite{choy20194d} as the model in our 3D branch to extract 3D features from the 3D point clouds.
Our 3D model is trained for 200 epochs with a batch size of 8.
Adam optimizer \cite{kingma2014adam} is used with a learning rate of 0.0001 and polynomial learning rate policy is used as the learning rate scheduler with power 0.9.
During inference, text-embeddings are computed by the ViT-L/14 CLIP model~\cite{clip} for each of the semantic categories and grounding queries.
We use the same pre-processing step and pre-trained dataset as OpenScene  \cite{openscene} (OpenSeg \cite{openseg}) for a fair comparison.

\subsection{Quantitative Results}

\noindent\textbf{Evaluation on zero-shot 3D semantic segmentation.}
We first compare our method with the state-of-the-art open-vocabulary scene understanding models and fully-supervised 3D segmentation models.
We report the mIoU for Scannet, Matterport3D, Scannet200, and Replica in Table~\ref{tab:main_results}. We find that our method achieves better results than the state-of-the-art open-vocabulary models and their variants on all the benchmarks. Besides, although our zero-shot model has noticeable performance drop compared with fully-supervised model, the gap of tail categories between the proposed method and those methods are relatively small (\eg 6.9 \vs 7.9 from CSC-Pretrain) on Scannet200. This demonstrates the strong potential of the proposed method for long-trailed 3D semantic segmentation tasks.

\noindent\textbf{Generalization to unseen dataset.}
To test the generalization ability of our proposed model, we evaluate it on an unseen dataset Replica \cite{straub2019replica} and report the results in Table~\ref{tab:main_results}.
The results shown that our proposed method significantly outperforms the state-of-the-art models on head, tail and all categories in Replica.
This demonstrates the strong generalization ability of the proposed model on novel datasets.

\noindent\textbf{Effectiveness of Different Distillation Settings.}
We compare our mask-based distillation method with point-based ones under different settings and report the performance of the 3D branch on Replica \cite{straub2019replica} in Table~\ref{tab:distillation}.
The supervisions for point-based method include:
(1) Fine-tuned CLIP feature, which follows the same setting as OpenScene \cite{openscene};
(2) Frozen diffusion feature extracted from the last layer of diffusion U-Net block.
We observe that distilling frozen diffusion features does not converge.
%
% This is due to the reason that the frozen diffusion features are designed to generate realistic images with rich details.
%
% They contain too much detailed information which are noise for semantic segmentation task.
%
%
Our proposed method, by introducing the semantic meaningful mask embedding output from the 2D branch as a fixed classifier, significantly boost the performance of the 3D branch.

\begin{table}[t]
\small
\centering
\caption{\textbf{Effectiveness of Different Distillation Settings.} We report mIoU of different methods on the Replica \cite{straub2019replica} dataset.}
\begin{tabular}{l|c|ccc}
% \small
    \toprule
    Setting              & Distillation Type   &  Head &Tail& All \\
    \midrule
    fine-tuned CLIP feature \cite{openscene} & Point-based    &  32.6 & 7.7  & 11.1          \\
    frozen diffusion feature      & Point-based         &  \multicolumn{3}{c}{ Divergence }        \\
    multimodal mask distillation (ours)    & Mask-based      & \textbf{43.3} & \textbf{8.0}   & \textbf{12.8}      \\
    \bottomrule
\end{tabular}
\label{tab:distillation}
\end{table}

\noindent\textbf{Ablation studies.}
We conduct ablation studies using the Replica dataset \cite{straub2019replica} and show the results in Table~\ref{tab:abl}.
We first analyze the effectiveness of combining 2D and 3D masks using equation~\ref{eq:probability}.
We observe that compared with using salient or geometric mask only, using both types of masks achieves the best performance.
This is intuitive as both salient patterns and geometric information are helpful to segment accurate class boundaries. 
We then analyze the effectiveness of different semantic features.
We find that discriminative and diffusion features serve as strong complementary to each other.
We also observe that using those two types of semantic features jointly can significantly outperform using any of them alone. 
\begin{table}[t]
\caption{\textbf{Performance of different model ablations.} We observe that each component of our model gains consistent improvements.}
      \centering
      \begin{tabular}{l|c}
\toprule
Method & mIoU\\
\midrule
Our full model & 17.5 \\
\hline
Without 2D (salient) mask & 12.8 \\ 
% \hline
Without 3D (geometric) mask & 16.5 \\ 
% \hline
Without discriminative (CLIP) features & 15.5 \\ 
% \hline
Without generative (Stable Diffusion) features & 15.3 \\ 
\bottomrule
\end{tabular}
\label{tab:abl}
\end{table}
% 20 classes bin

\begin{figure*}[t!]
\centering
\setlength{\tabcolsep}{0.1em}
\begin{tabular}{>{\centering\scriptsize\arraybackslash}m{0.12\linewidth}m{0.198\linewidth}m{0.198\linewidth}m{0.198\linewidth}m{0.198\linewidth}}
Input 3D&
\includegraphics[width=0.9\linewidth]{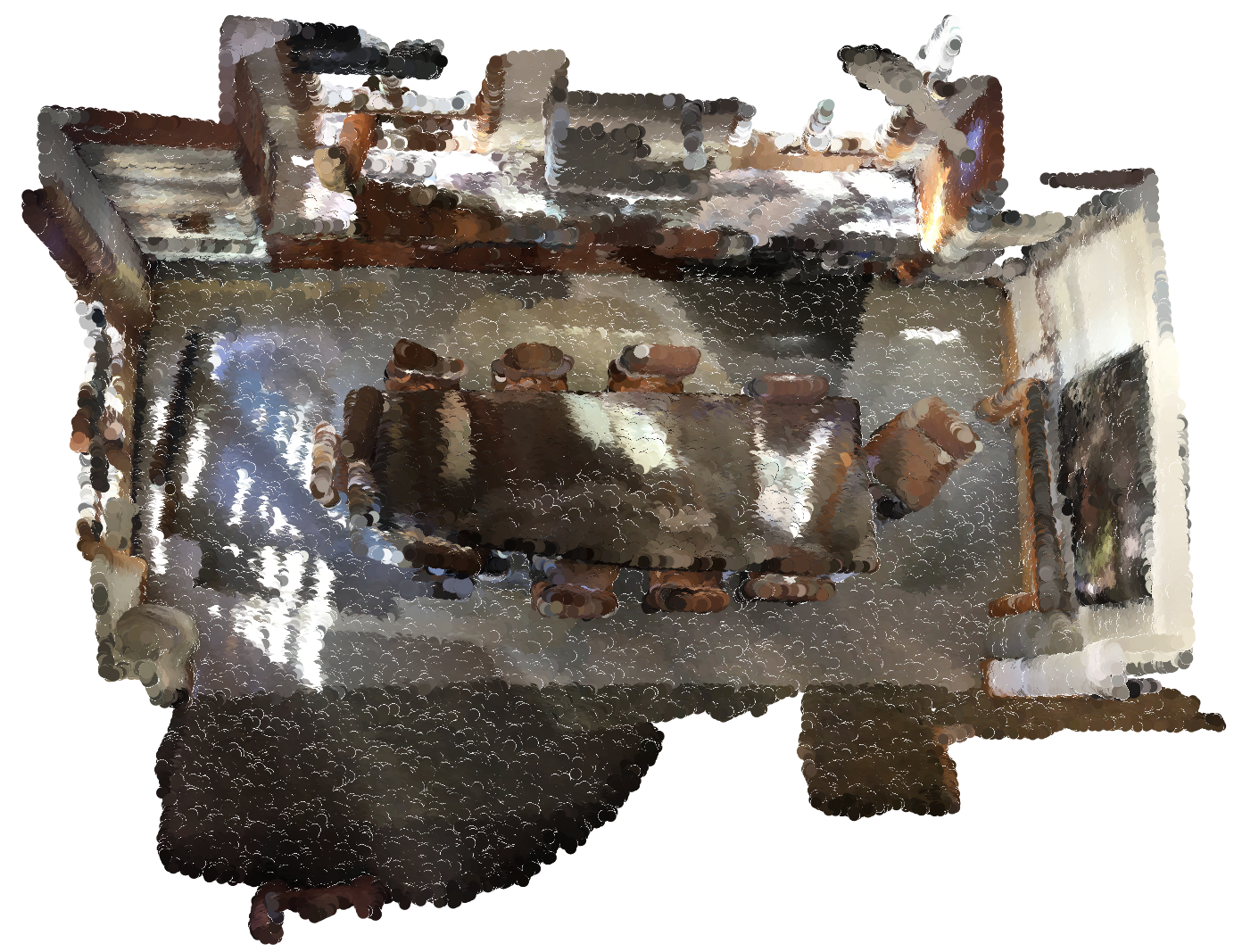} &
\includegraphics[width=0.9\linewidth]{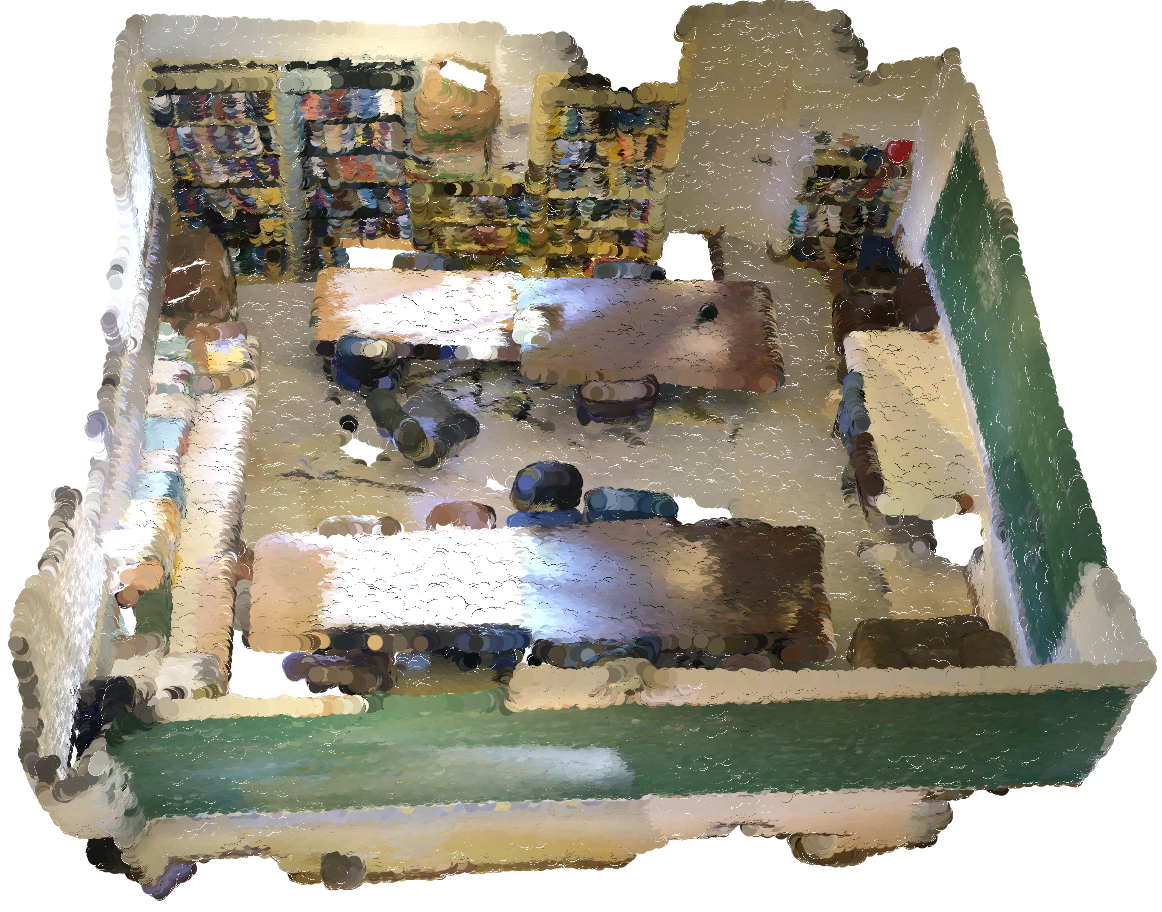} &
\includegraphics[width=0.9\linewidth]{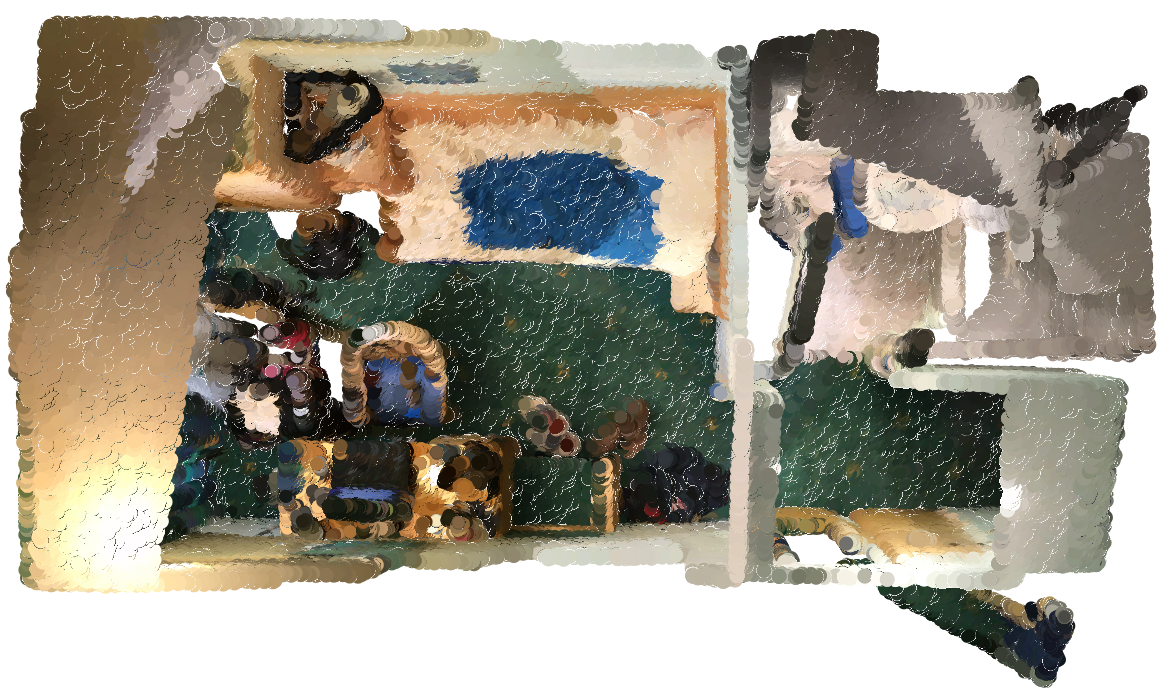} &
\includegraphics[width=0.9\linewidth]{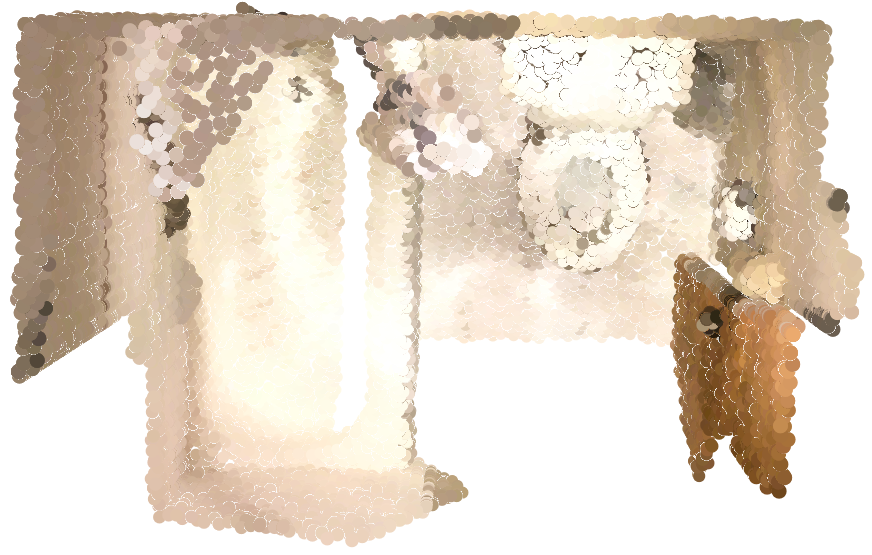} \\
Ground Truth &
\includegraphics[width=0.9\linewidth]{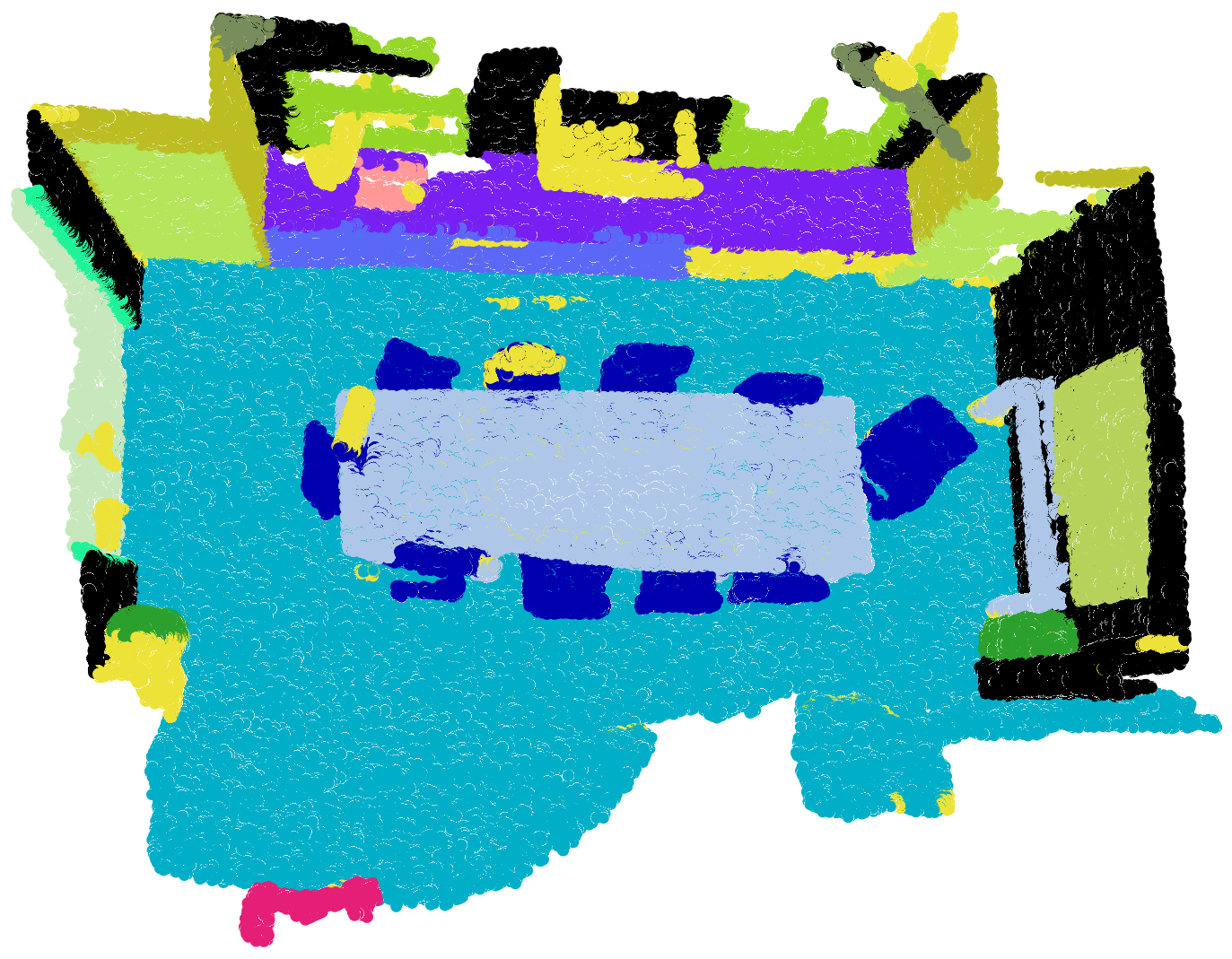} &
\includegraphics[width=0.9\linewidth]{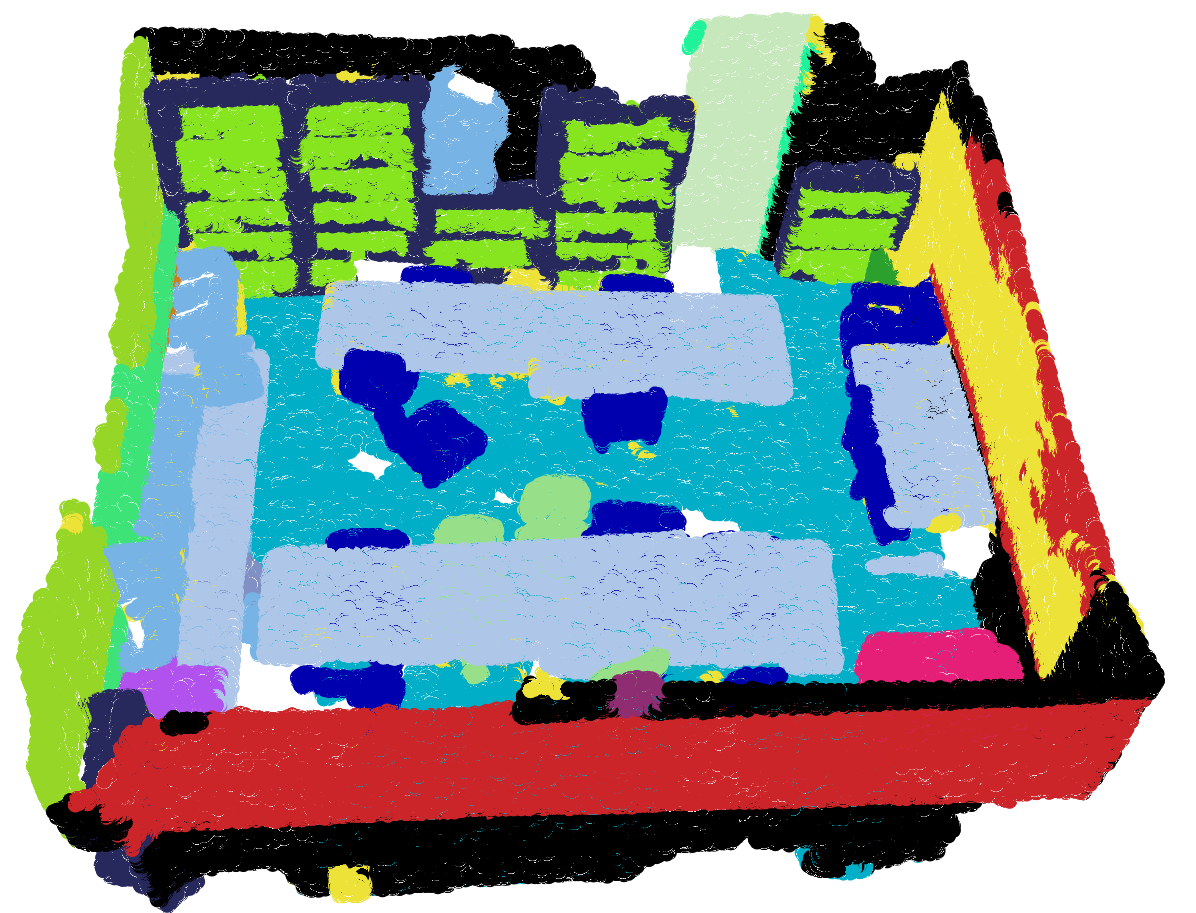} &
\includegraphics[width=0.9\linewidth]{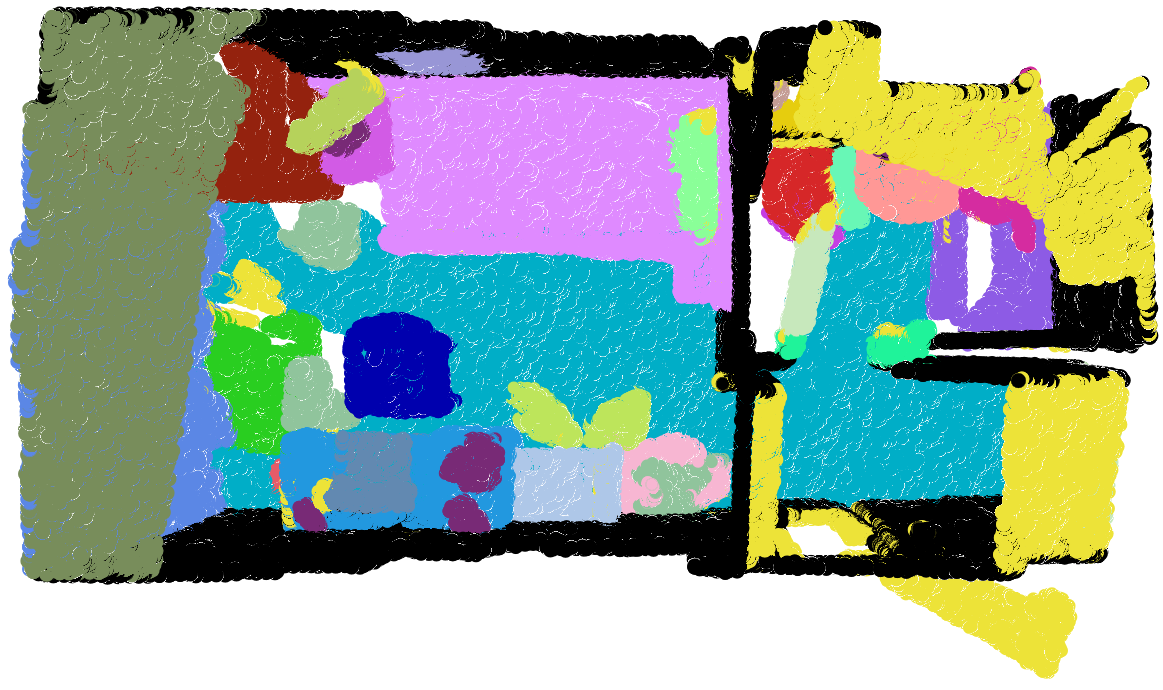} &
\includegraphics[width=0.9\linewidth]{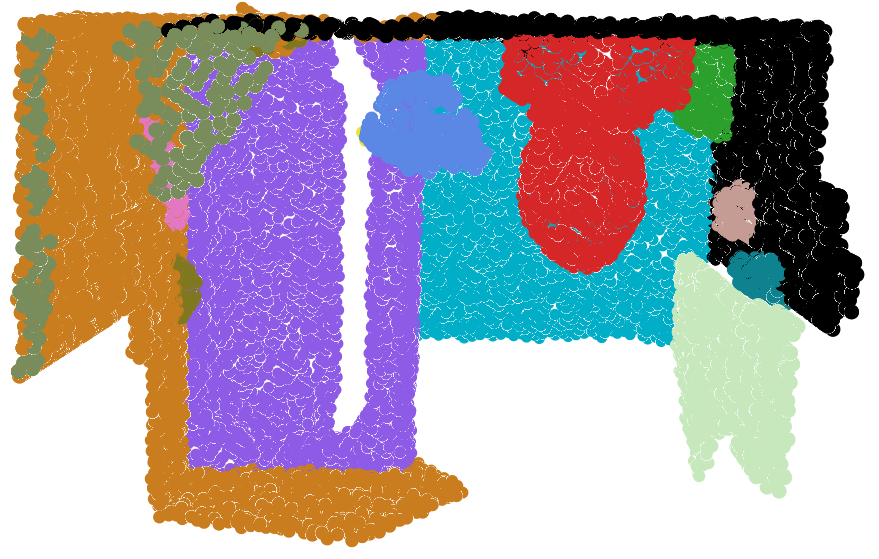} \\
OpenScene  &
\includegraphics[width=0.9\linewidth]{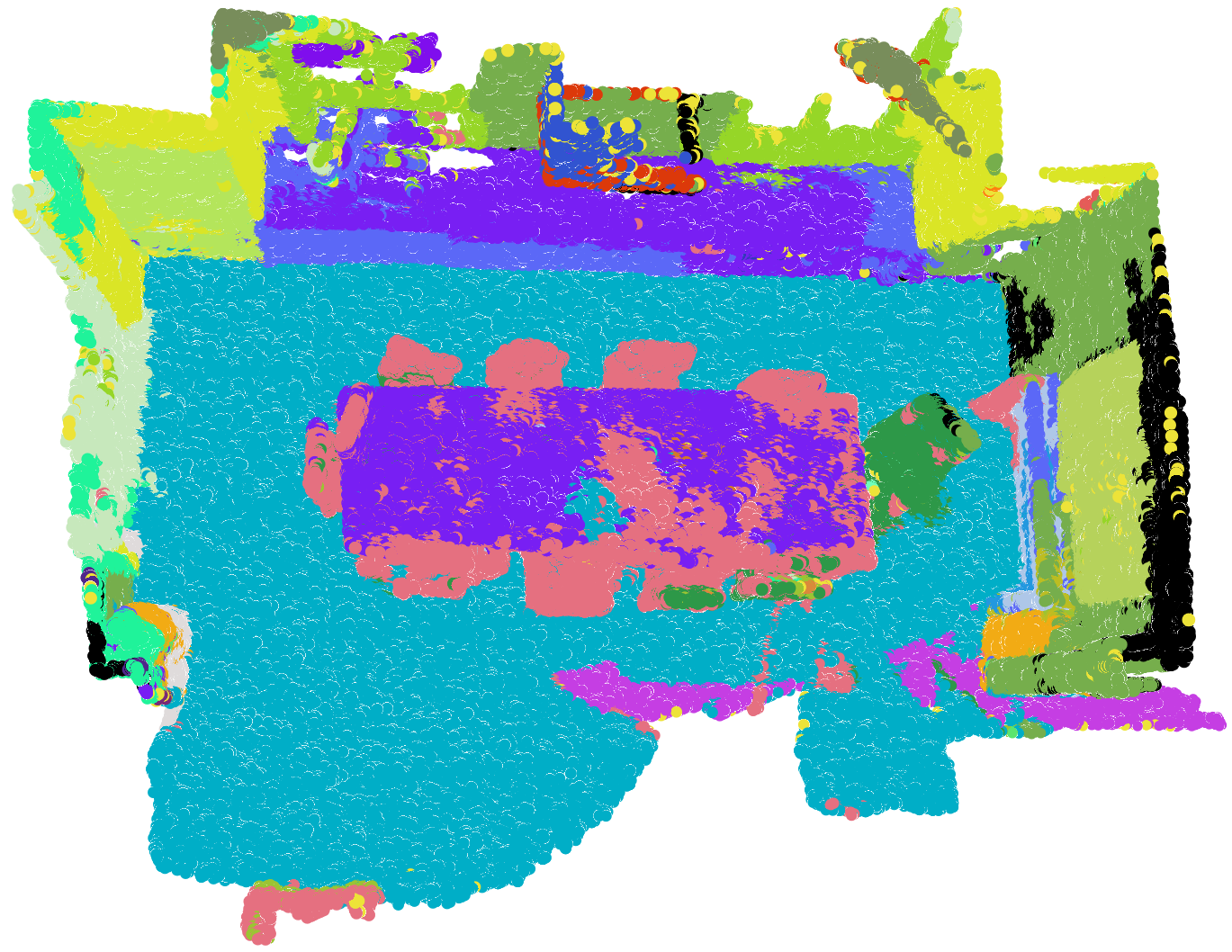} &
\includegraphics[width=0.9\linewidth]{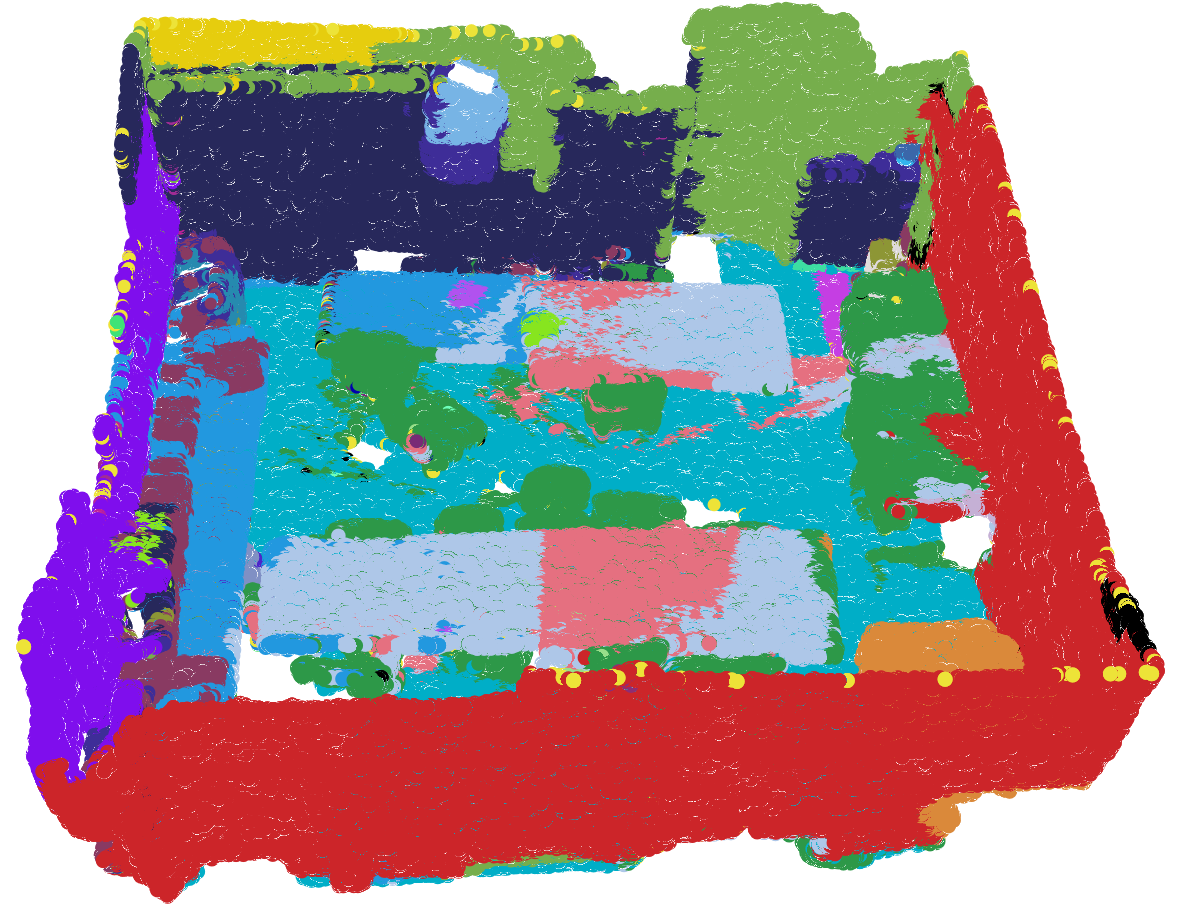} &
\includegraphics[width=0.9\linewidth]{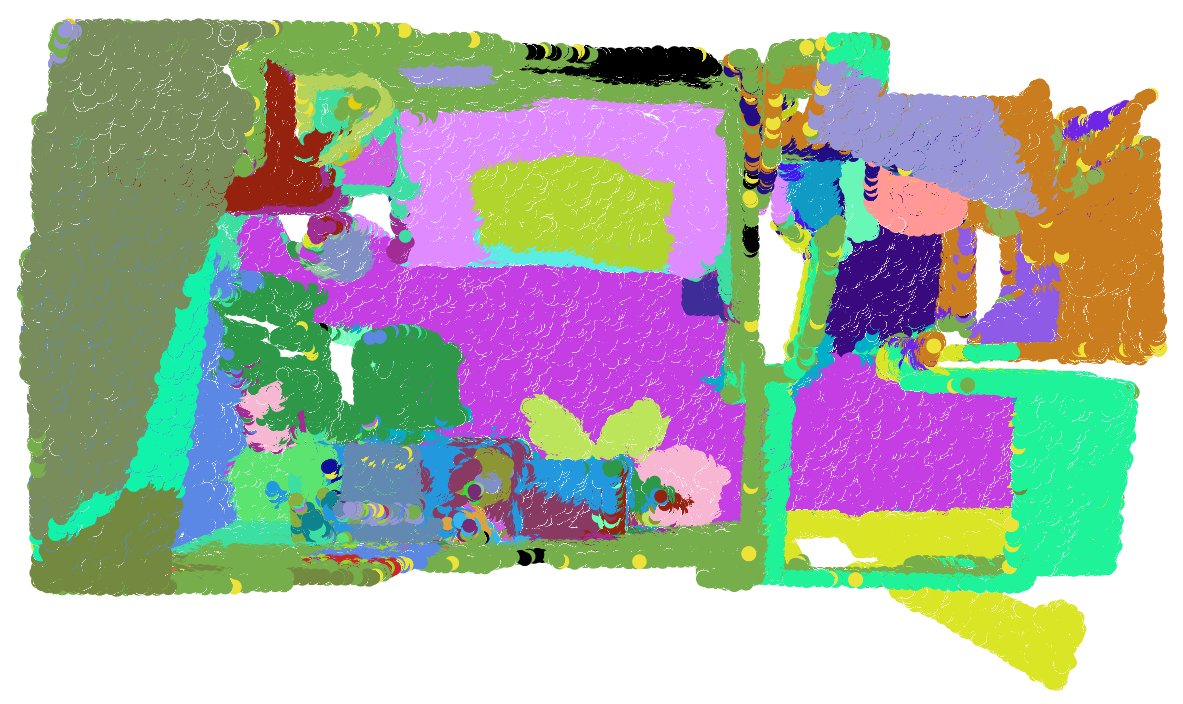} &
\includegraphics[width=0.9\linewidth]{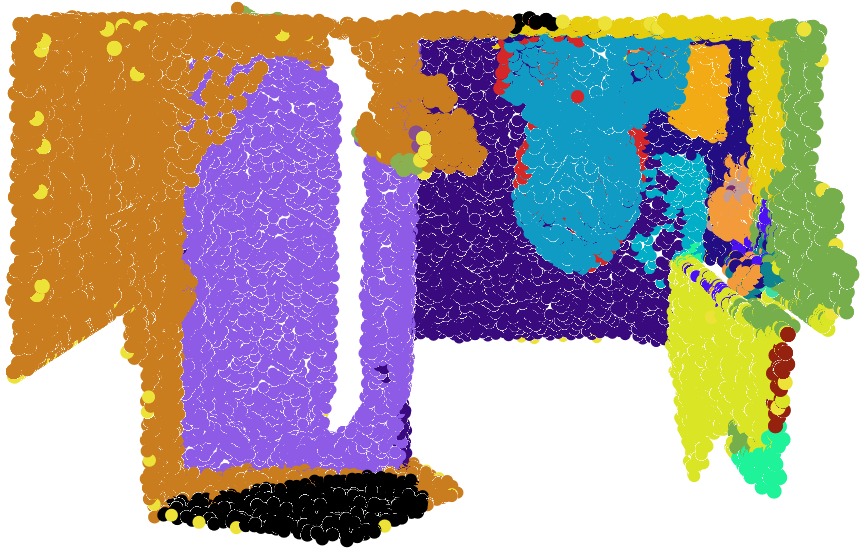} \\
Ours  &
\includegraphics[width=0.9\linewidth]{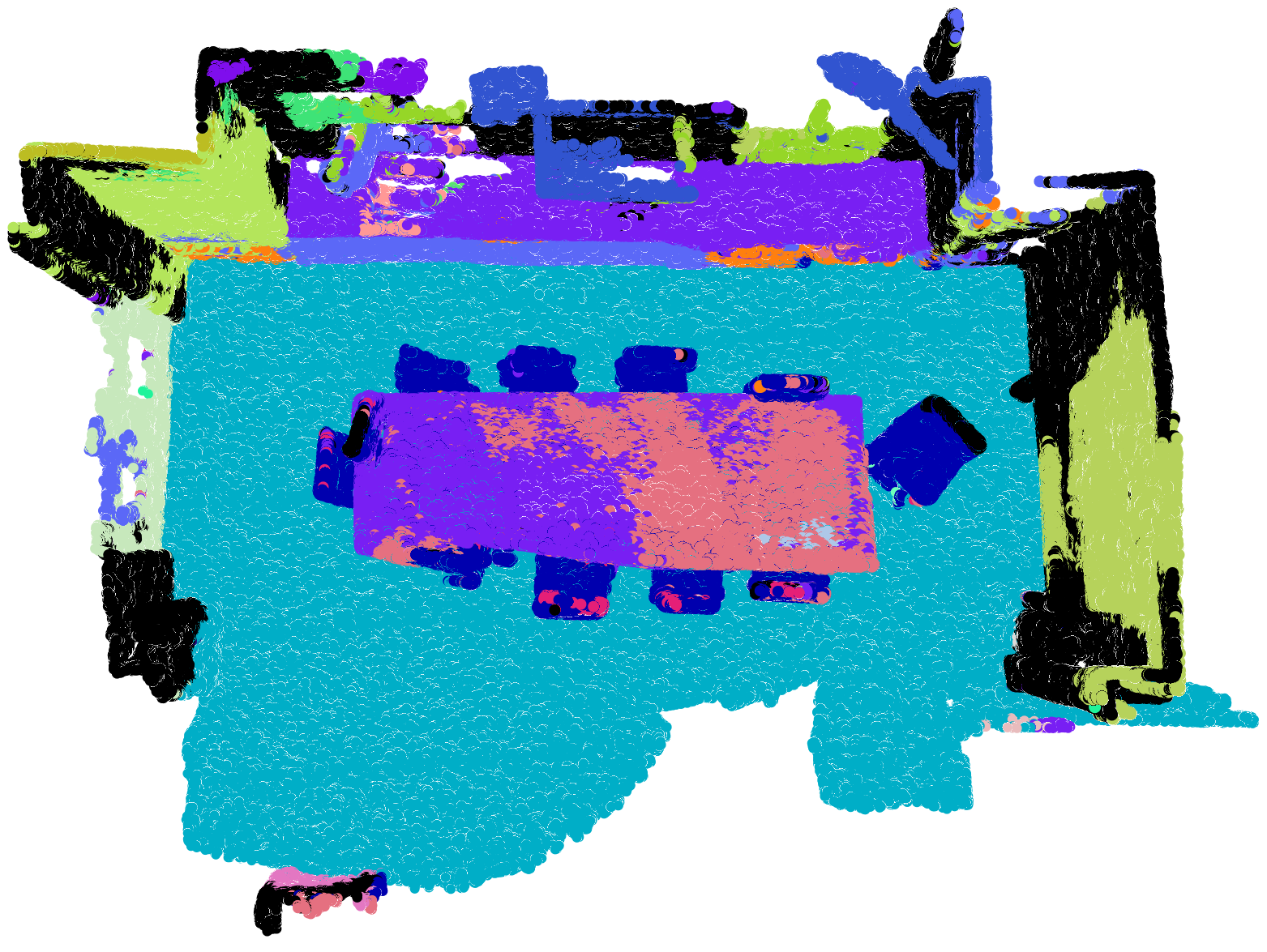} &
\includegraphics[width=0.9\linewidth]{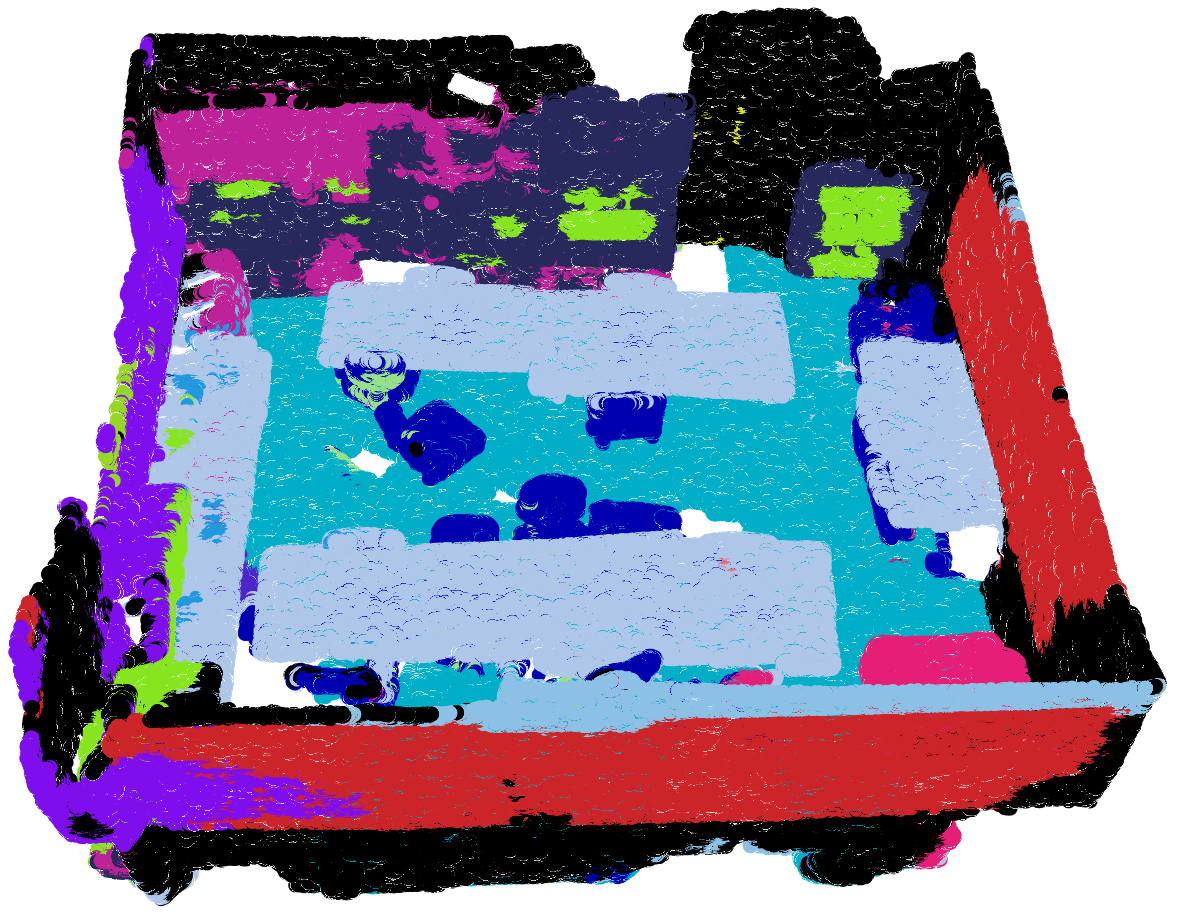} &
\includegraphics[width=0.9\linewidth]{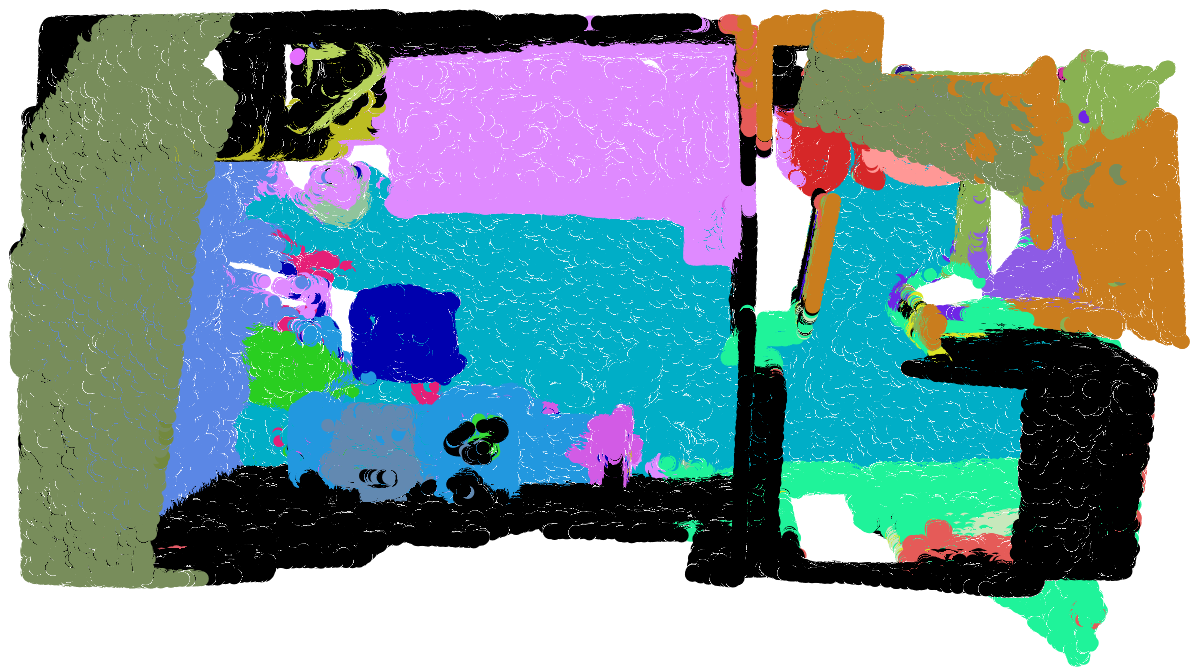} &
\includegraphics[width=0.9\linewidth]{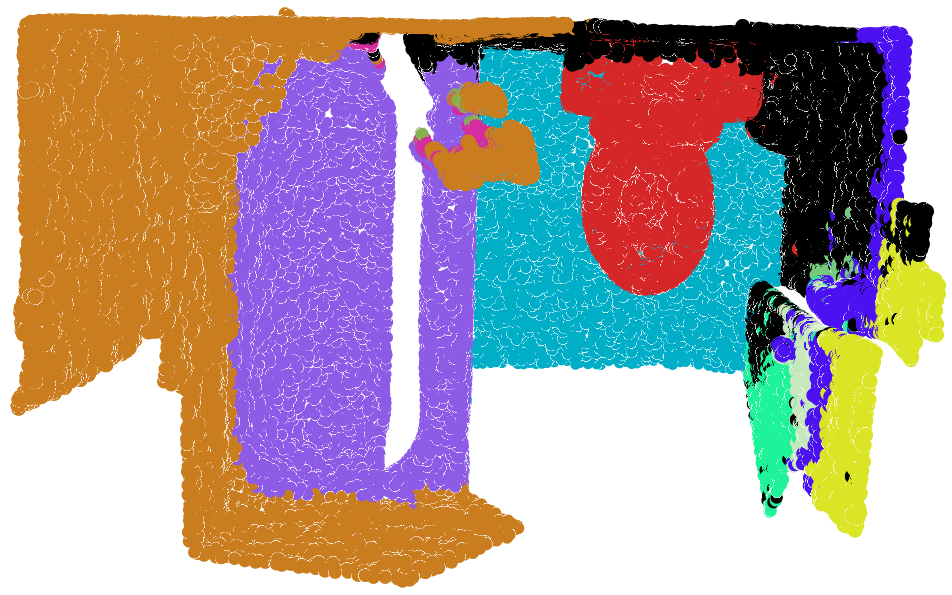} \\
\end{tabular}
\includegraphics[width=1.0\linewidth]{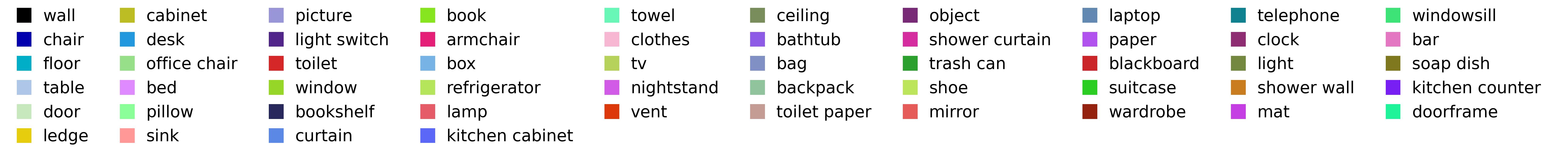} \\
\caption{\textbf{Qualitative results from our model and OpenScene on zero-shot semantic segmentation.} We visualize the segmentation results on the validation set of ScanNet200 \cite{rozenberszki2022language}. We observe that our model can predict coherent masks with accurate semantic labels compared to OpenScene for both head and tail categories.}
\label{fig:scannet200}
\end{figure*}

\begin{figure*}[t!]
    \centering
    \setlength{\tabcolsep}{0.2em}
    \begin{tabular}{>{\centering\scriptsize\arraybackslash}m{0.1\linewidth}>{\centering\scriptsize\arraybackslash}m{0.197\linewidth}>{\centering\scriptsize\arraybackslash}m{0.21\linewidth}>{\centering\scriptsize\arraybackslash}m{0.21\linewidth}>{\centering\scriptsize\arraybackslash}m{0.21\linewidth}}
    Input 3D &
    \includegraphics[width=\linewidth]{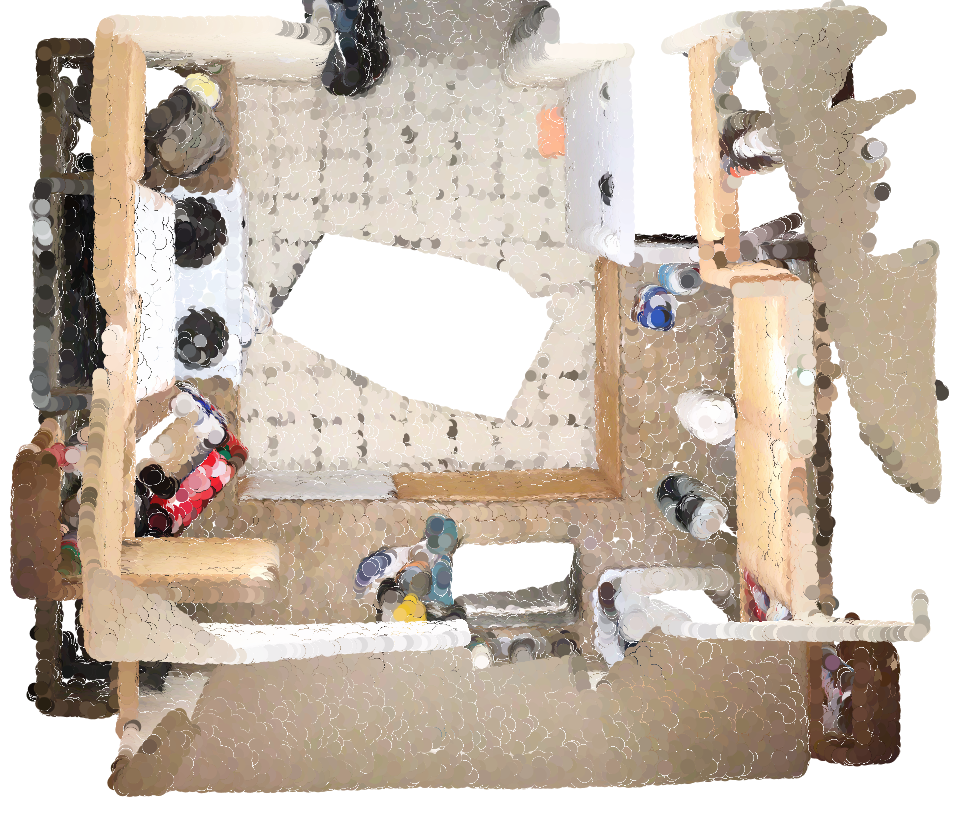} &
    \includegraphics[width=\linewidth]{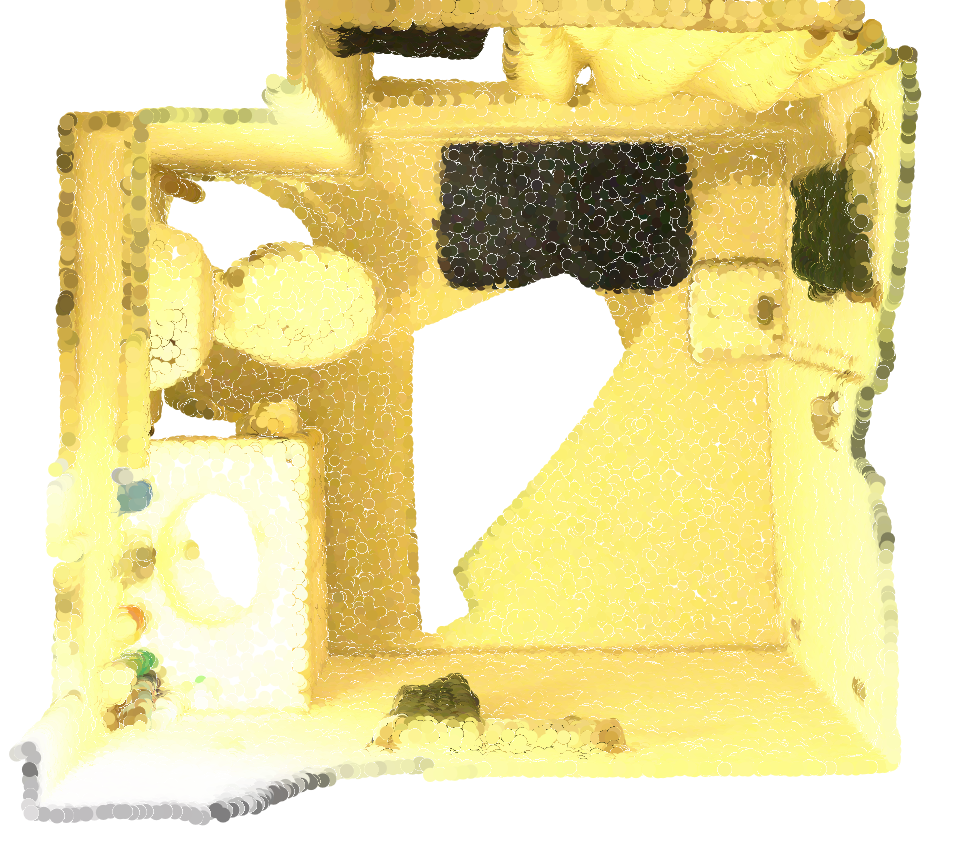} &
    \includegraphics[width=\linewidth]{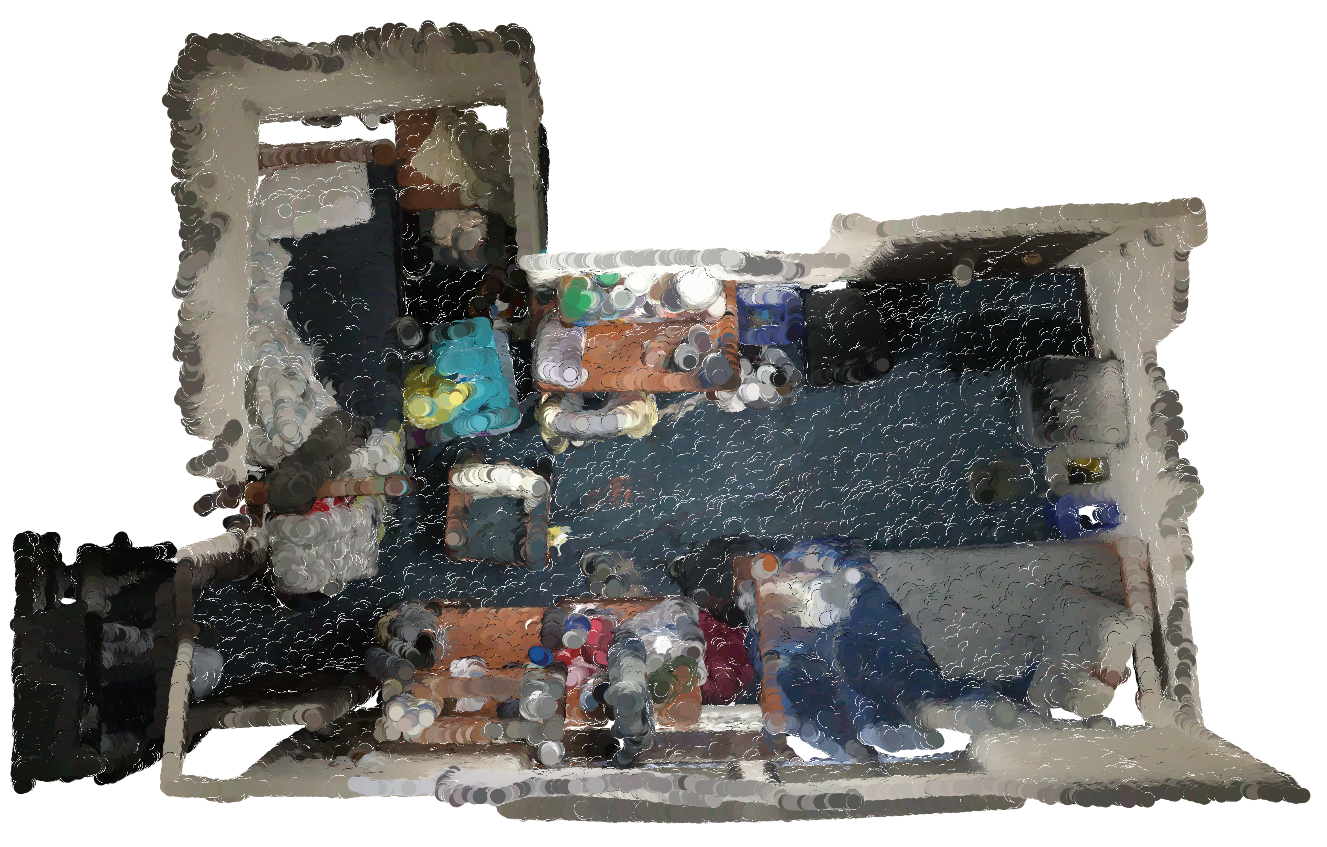} &
    \includegraphics[width=\linewidth]{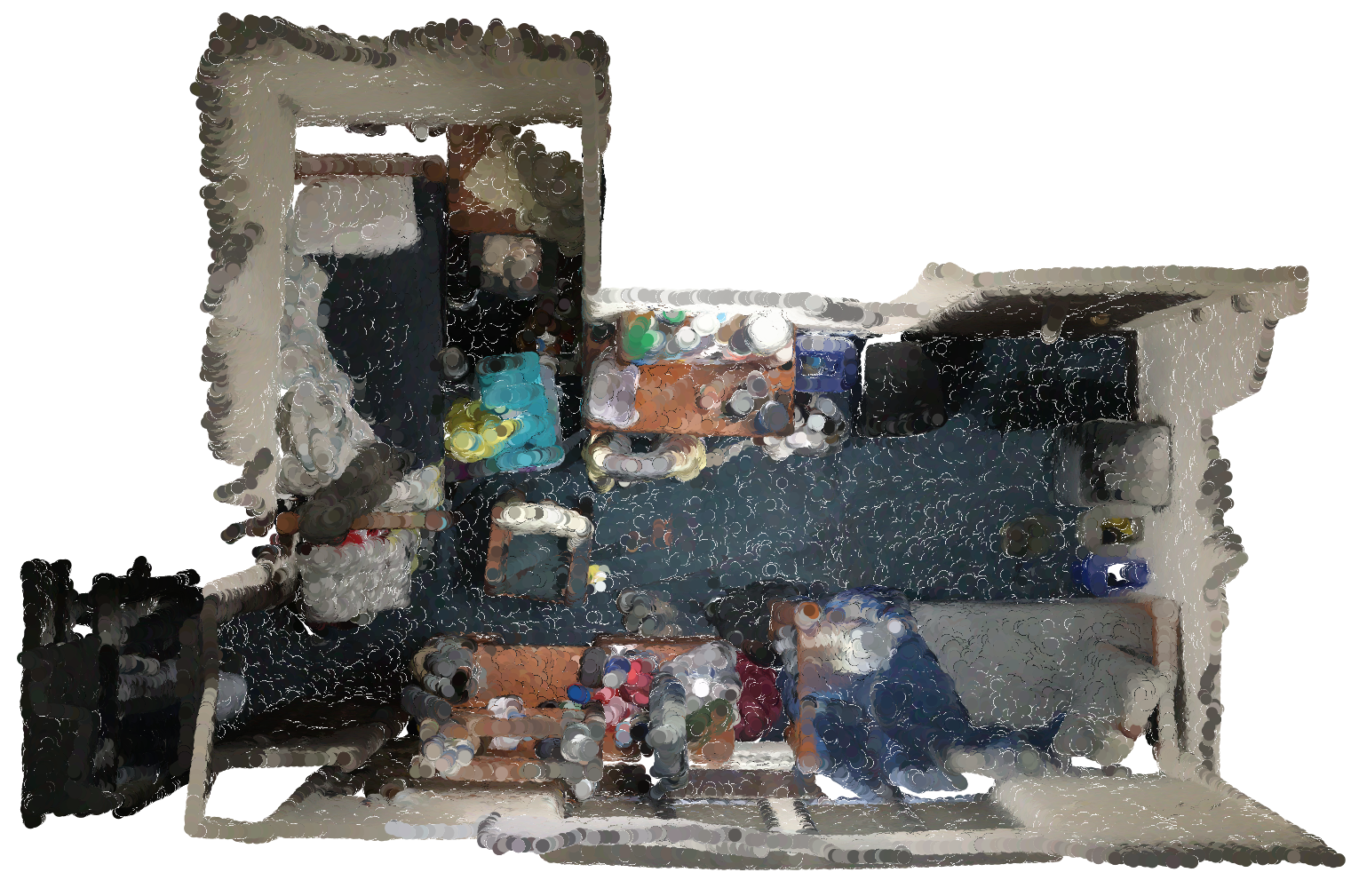} \\
    Ground Truth &
    \includegraphics[width=\linewidth]{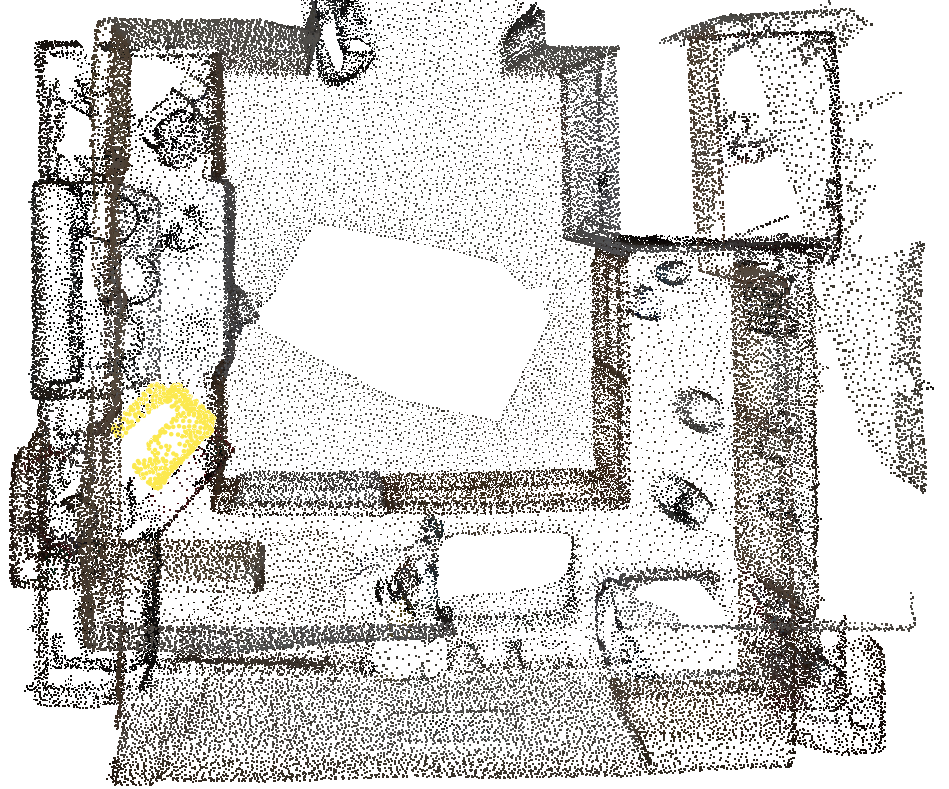} &
    \includegraphics[width=\linewidth]{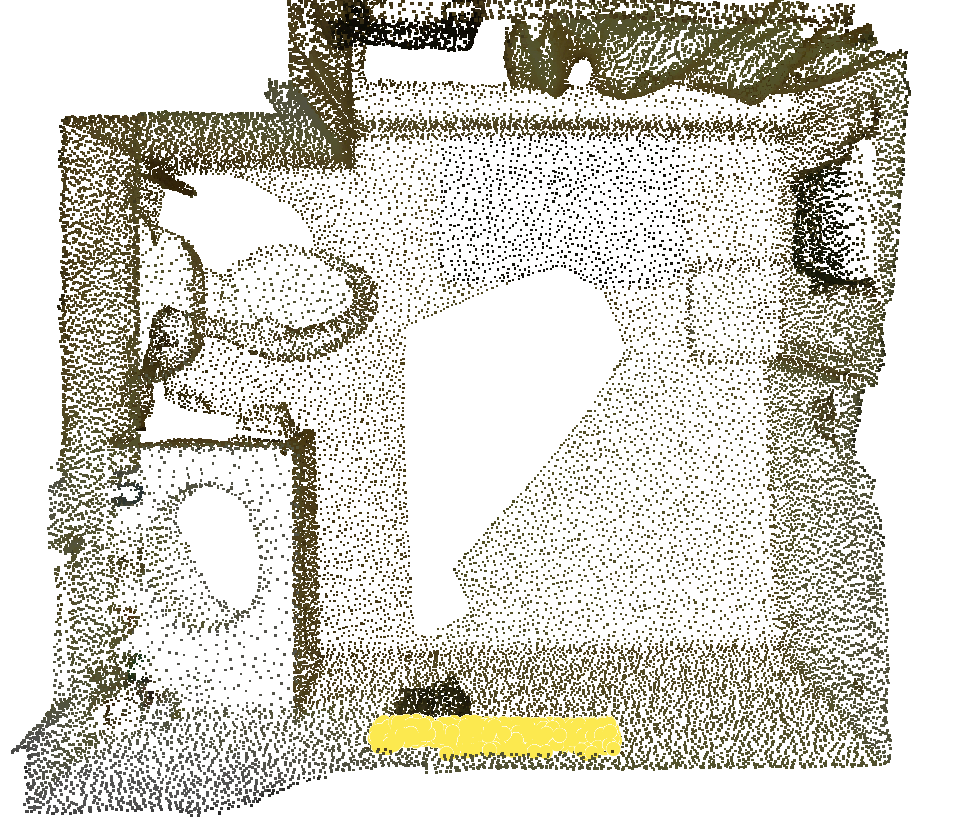} &
    \includegraphics[width=\linewidth]{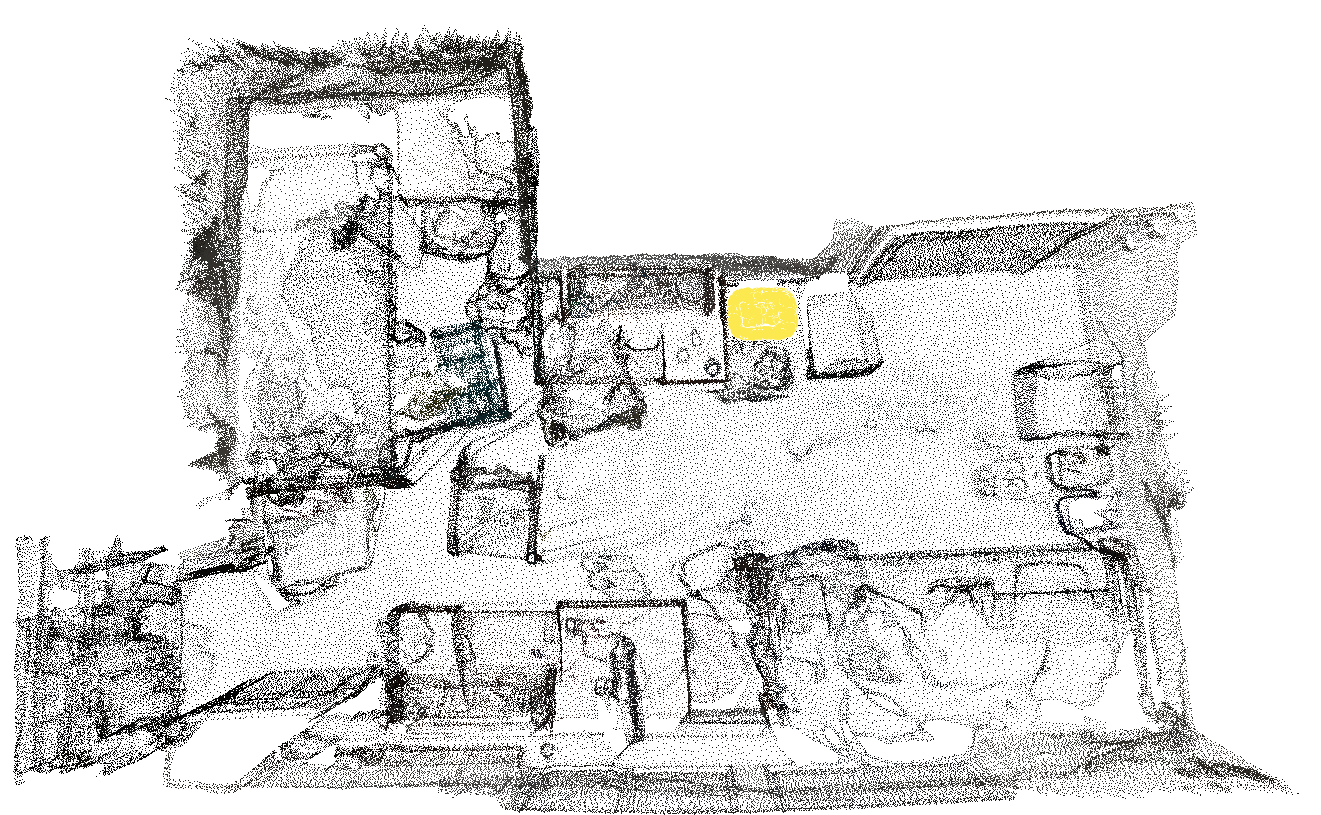} &
    \includegraphics[width=\linewidth]{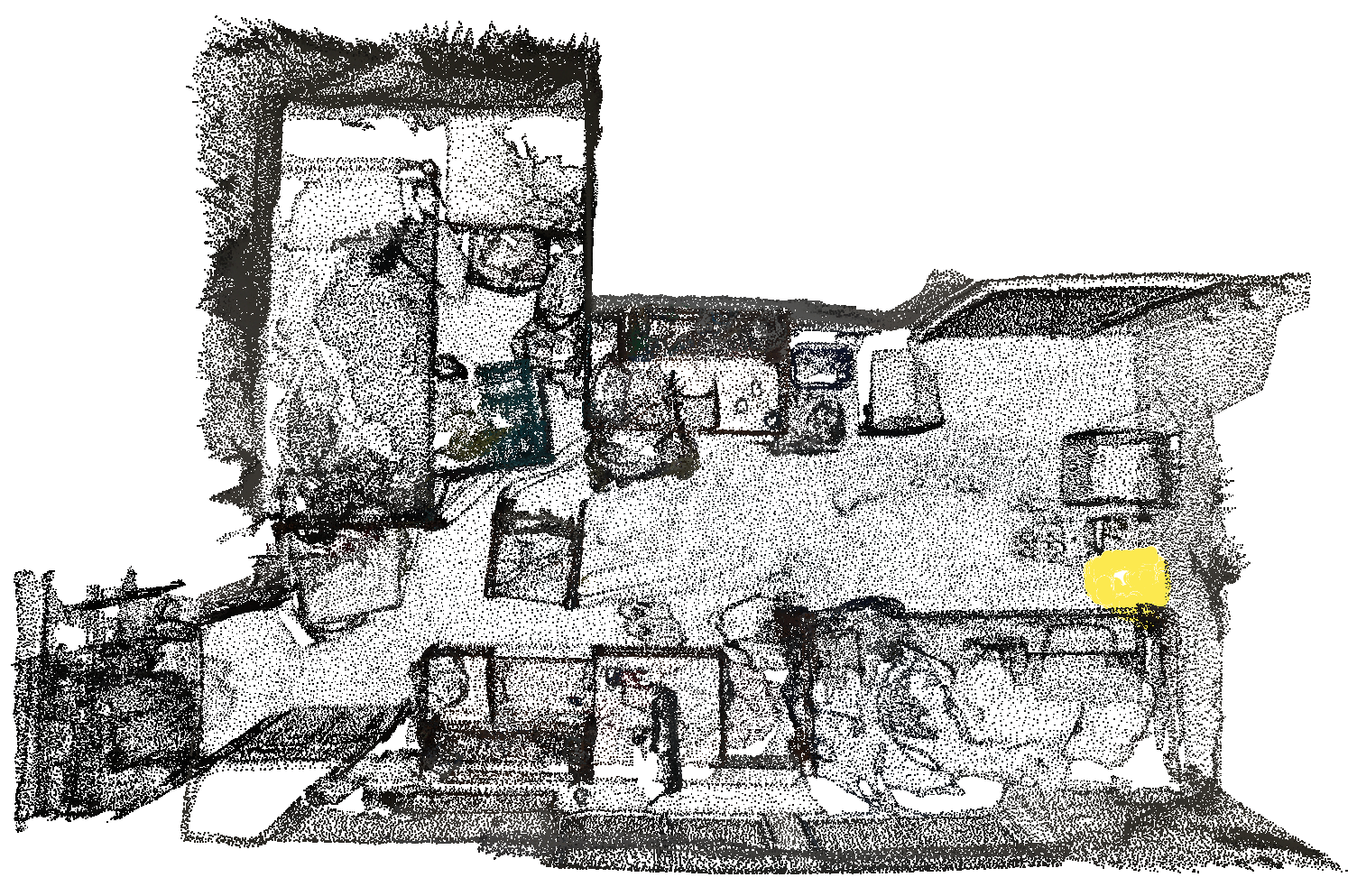} \\
    OpenScene &
    \includegraphics[width=\linewidth]{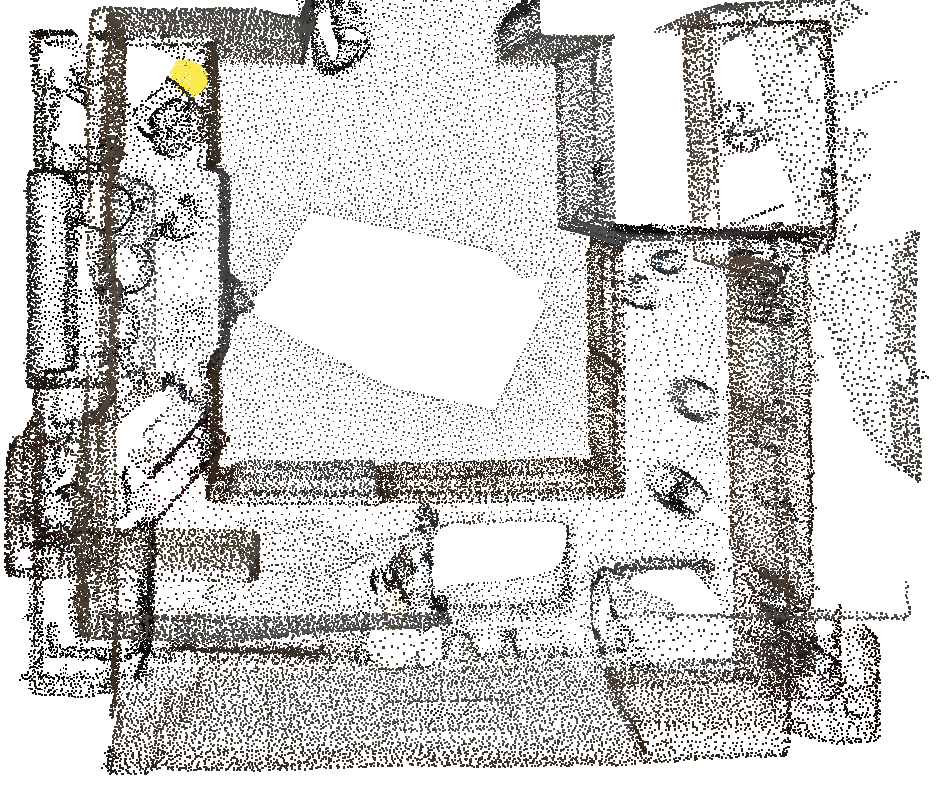} &
    \includegraphics[width=\linewidth]{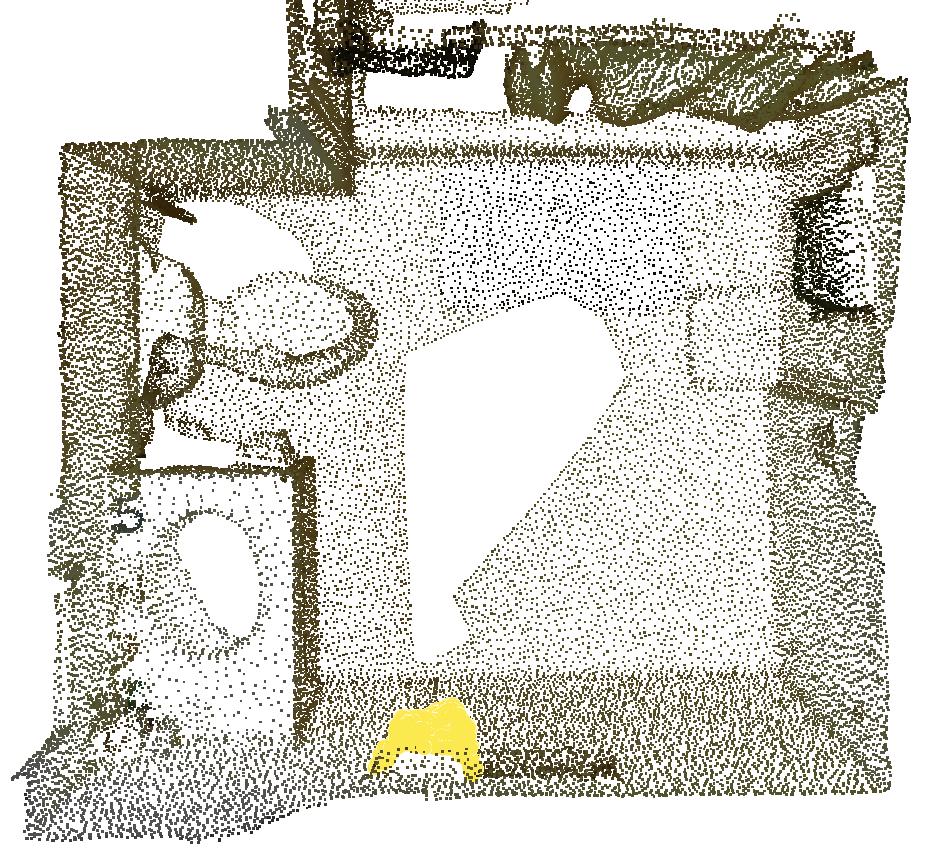} &
    \includegraphics[width=\linewidth]{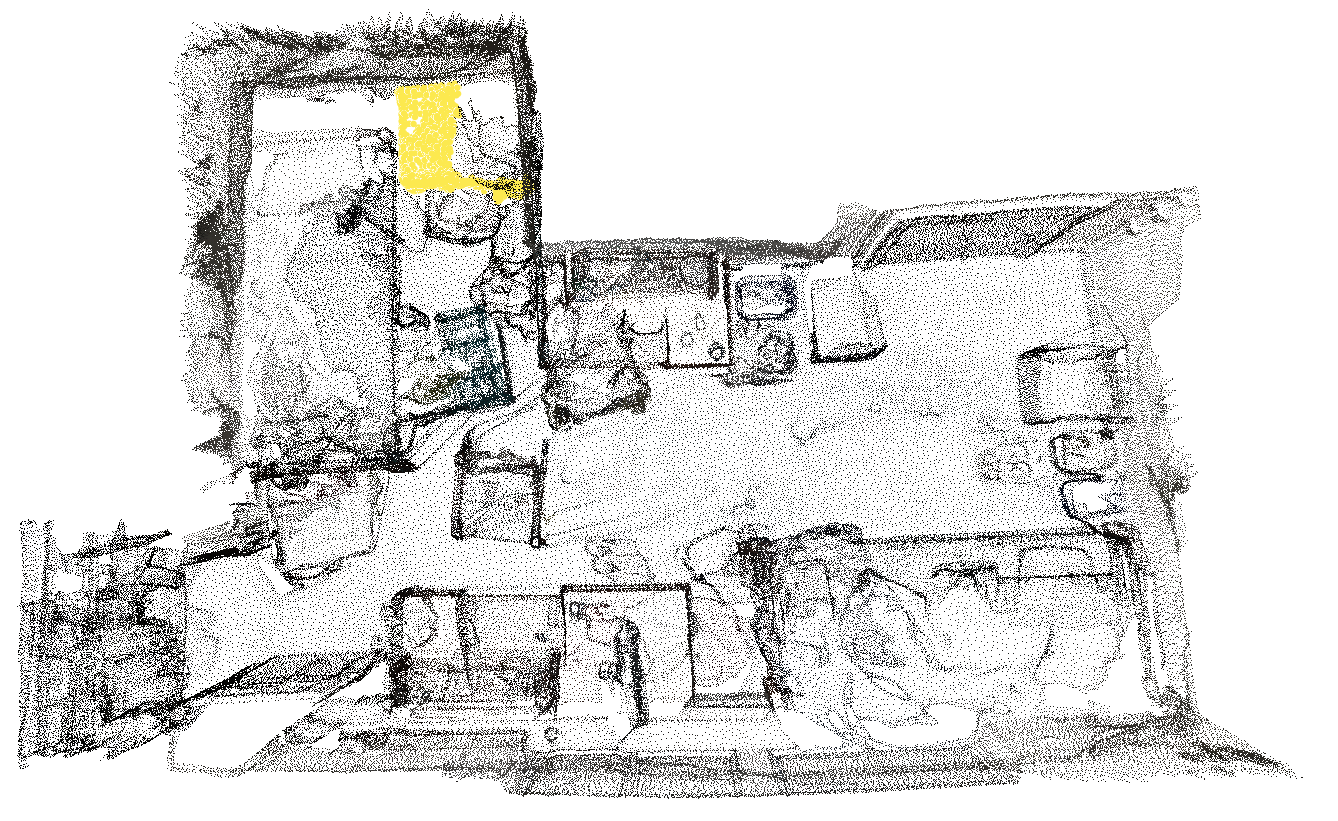} &
    \includegraphics[width=\linewidth]{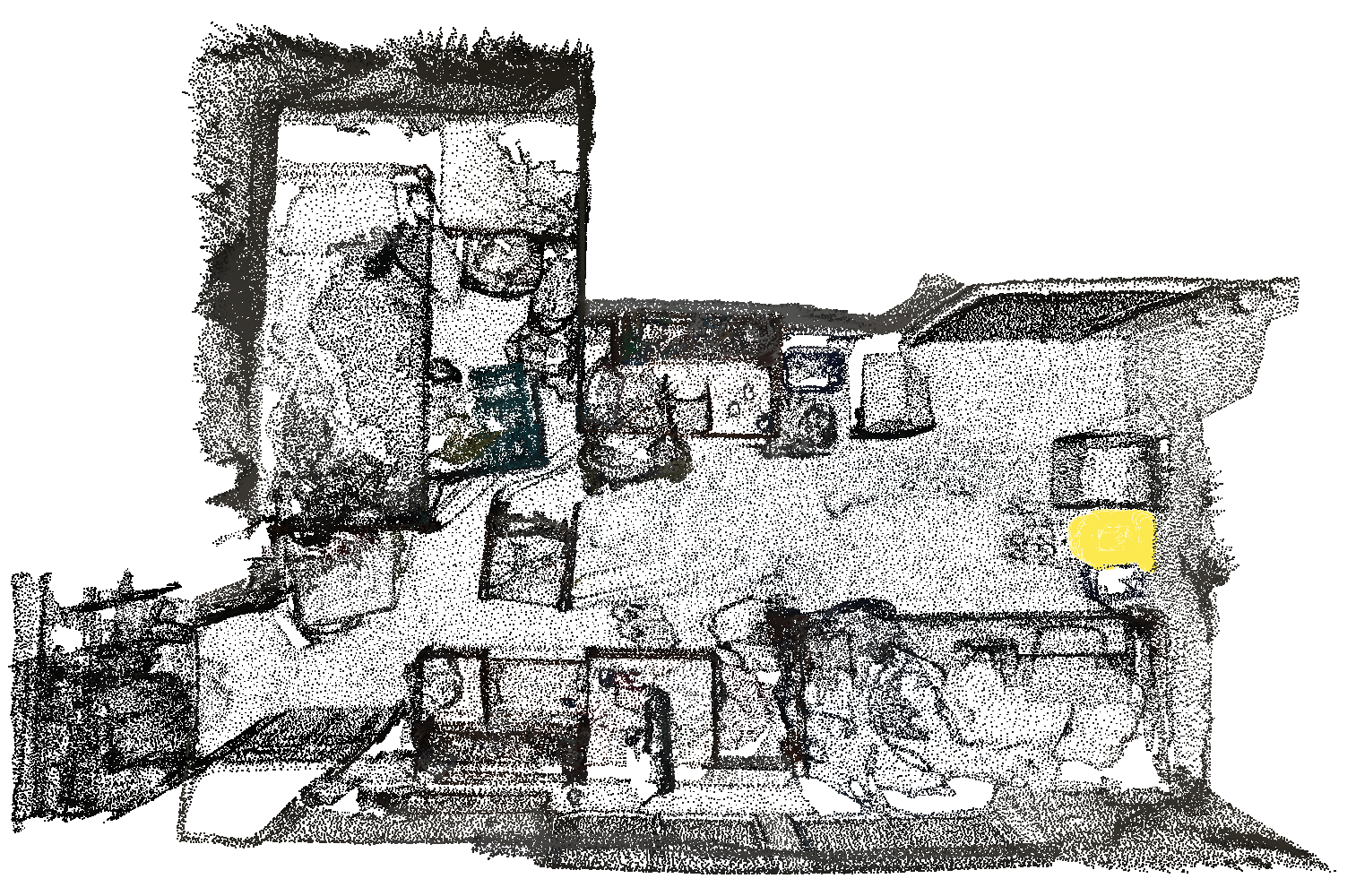} \\
    Ours  &
    \includegraphics[width=\linewidth]{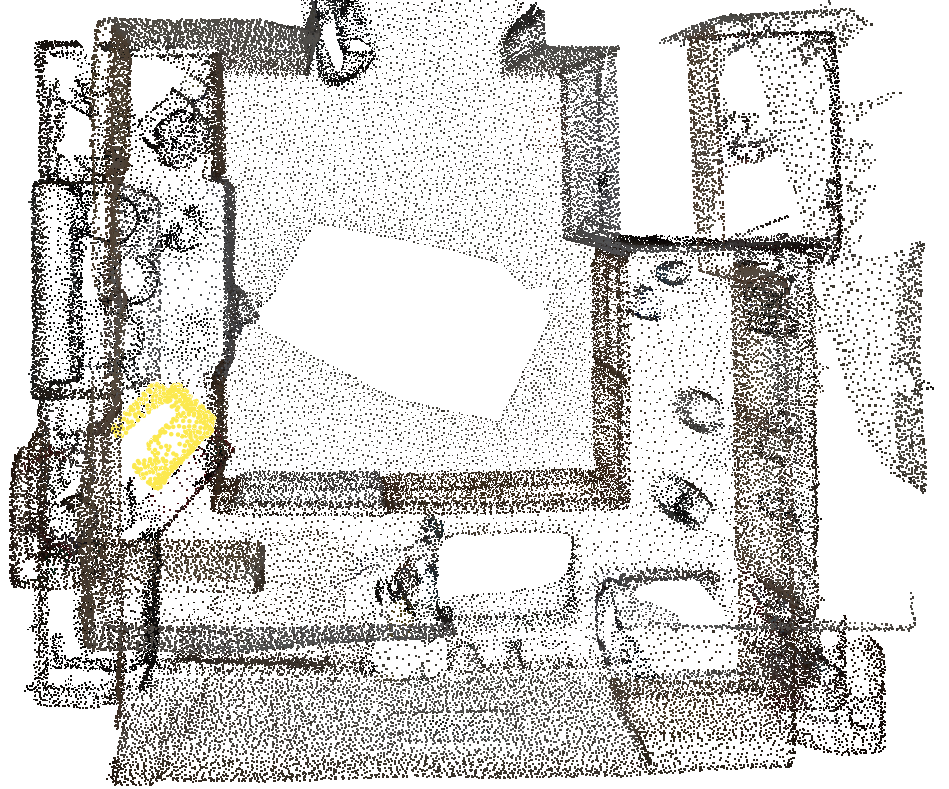} &
    \includegraphics[width=\linewidth]{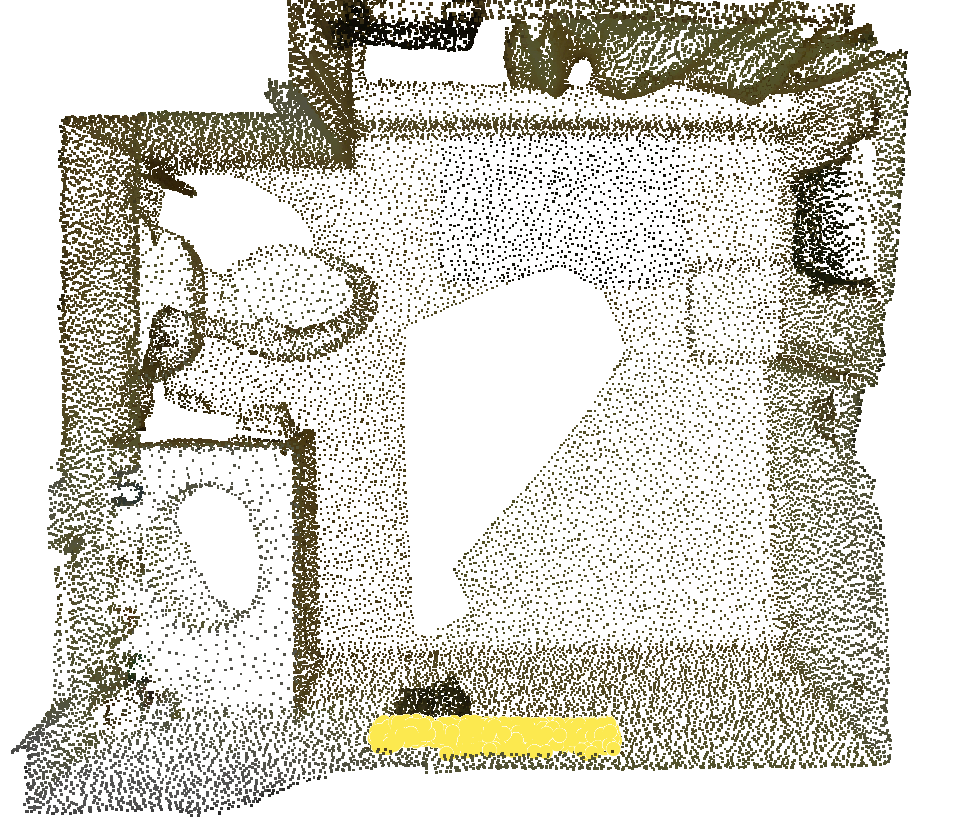} &
    \includegraphics[width=\linewidth]{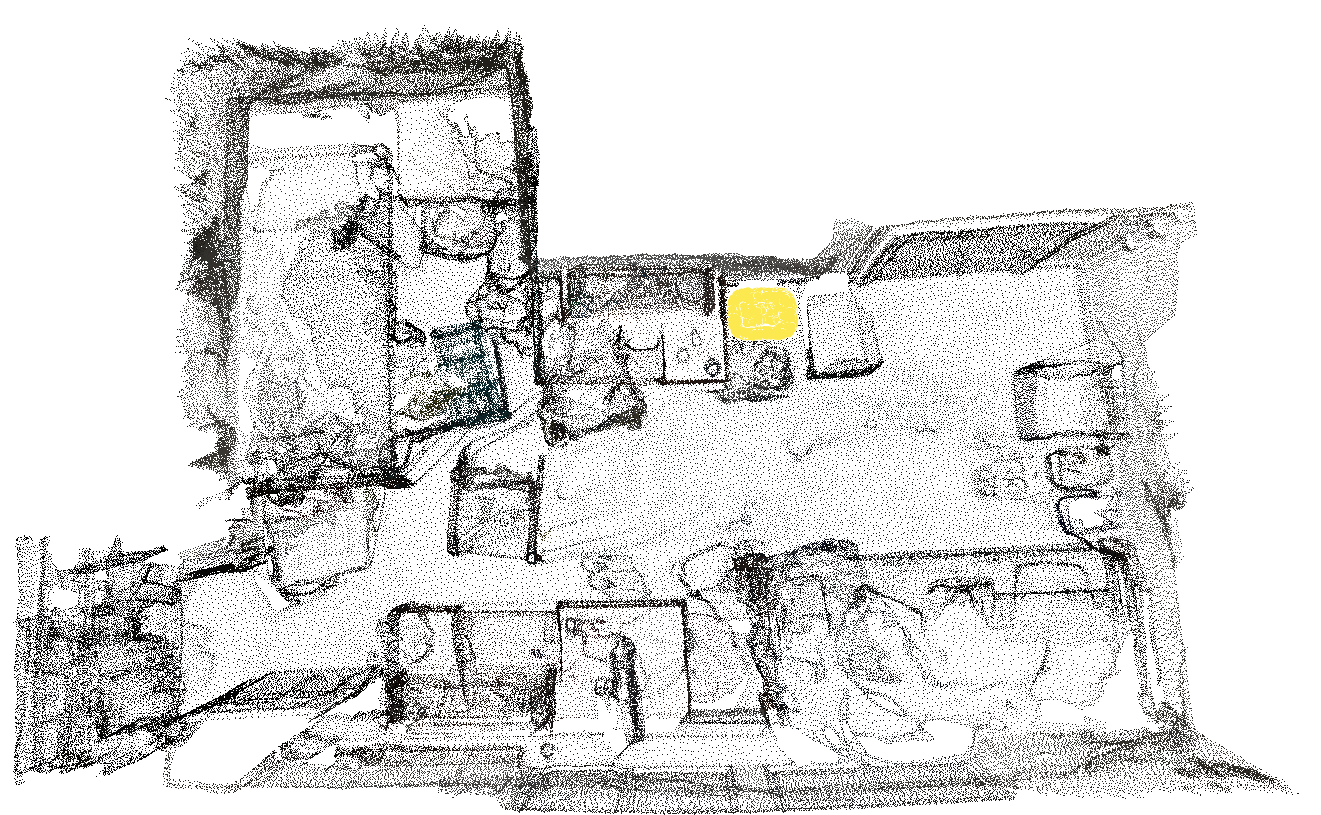} &
    \includegraphics[width=\linewidth]{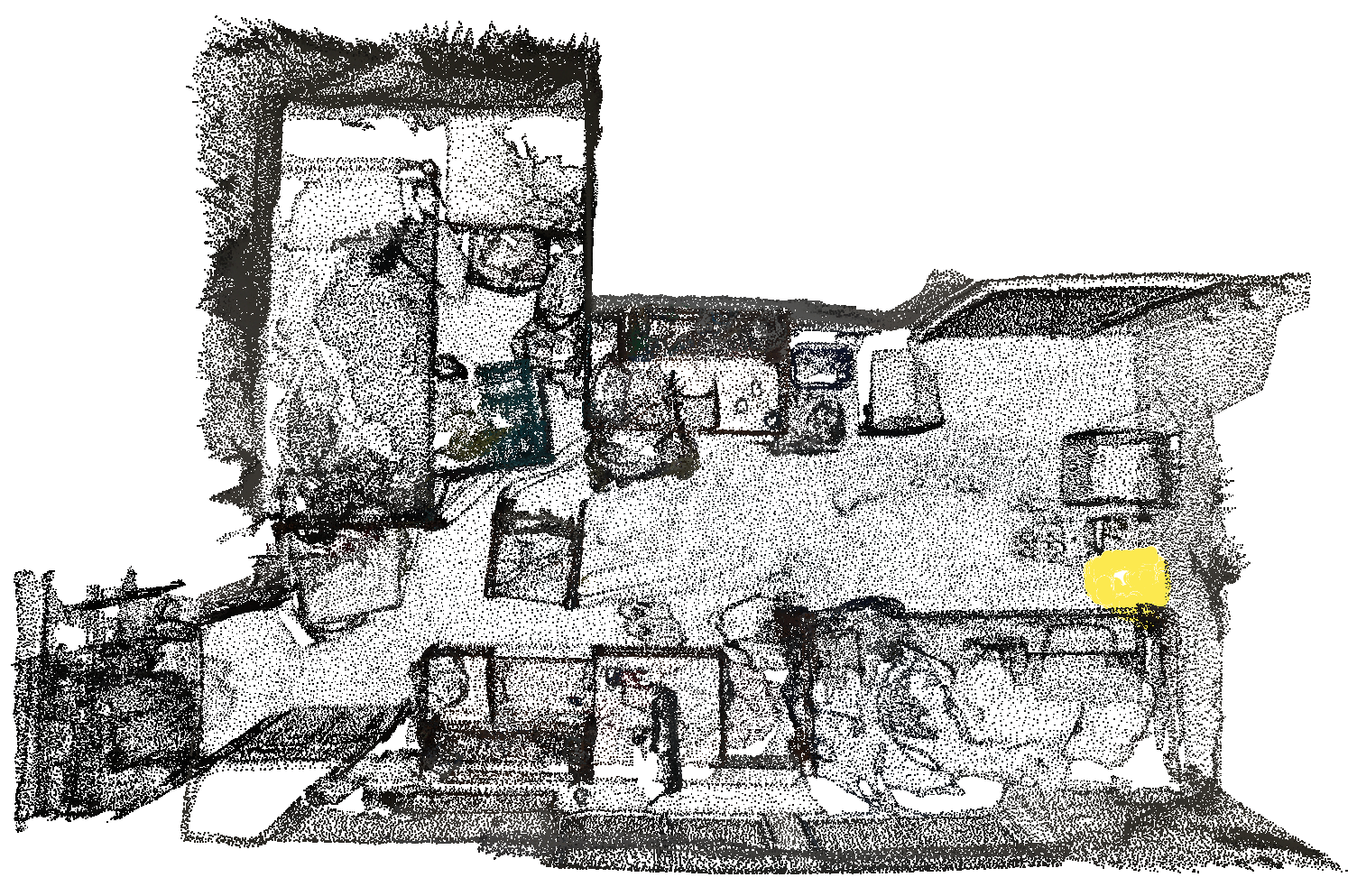} \\
    Text Query &
    \begin{tabular}{>{\centering\scriptsize}m{\linewidth}}
     ``It's the shorter \\ red box''
    \end{tabular}&
    \begin{tabular}{>{\centering\scriptsize}m{\linewidth}}
     ``The rack with the \\ smaller, more \\ wrinkled towel''
    \end{tabular}&
    \begin{tabular}{>{\centering\scriptsize}m{\linewidth}}
     ``It is the trash can \\ next to the desk''
    \end{tabular}&
    \begin{tabular}{>{\centering\scriptsize}m{\linewidth}}
     ``A blue waste \\ basket  possibly \\ for recycling''
    \end{tabular}\\
    &
    (a) &
    (b) &
    (c) &
    (d) 
    \end{tabular}
    % \vspace{0.4cm}
    \caption{\textbf{Qualitative results from our model and OpenScene on zero-shot visual grounding.} Our open-vocabulary semantic understanding model is capable of handling different types of novel and compositional queries. Novel object classes as well as objects described by colors, shapes, appearances, locations, and usages are successfully retrieved by our method. Note that the located points are colored in yellow.}
    \label{fig:visual_grounding}
    % \vspace{-3mm}
\end{figure*}

\subsection{Qualitative analysis}
\noindent\textbf{Visualizations of zero-shot semantic segmentation.}
In Fig.~\ref{fig:scannet200}, we provide qualitative analysis of our approach and OpenScene for the zero-shot 3D semantic
segmentation task. Compared with OpenScene, our model generates coherent and consistent masks (\eg, the table mask in first column and the bed mask in third column) thanks to the mask-instance representations. It predicts accurate semantic labels for both head and tail categories by leveraging both CLIP and diffusion features. 

\noindent\textbf{Visualizations of visual grounding results.}
We provide qualitative analysis of our approach and OpenScene for the zero-shot visual grounding task in Fig.~\ref{fig:visual_grounding}. We observe that our model can accurately identify objects given complicated text queries. It demonstrates that the proposed method, \emph{Diff2Scene}, has good capability at the following types of queries. Fig.~\ref{fig:visual_grounding} (a) describes object shape and color, and even in comparative degree (\emph{It's the shorter, red box}); Fig.~\ref{fig:visual_grounding} (b) describes a rare object (\emph{rack}) and its surrounded object with surface appearance descriptions (\emph{wrinkled towel}); Fig.~\ref{fig:visual_grounding} (c) describes the relative location of the object (\emph{next to the desk}); Fig.~\ref{fig:visual_grounding} (d) describes the usage of the object (\emph{recycling}). 
In addition, we can see that given vague usage descriptions without common category names like \emph{trash bin} in the text prompts, the model can still accurately identify the object.

\section{Conclusion}
In this paper, we investigate the problem of leveraging frozen representations from large text-to-image diffusion models for open-vocabulary 3D semantic understanding. \emph{Diff2Scene} sets a new state-of-the-art in the zero-shot 3D semantic segmentation task and shows promising performance in the visual grounding task. Our method also shows outstanding generalization ability towards unseen datasets and novel text queries. It provides a new way to effectively leverage generative text-to-image foundation models for 3D semantic scene understanding tasks.

There are several limitations of the proposed model.
First, while our model achieves better performance compared to existing methods in small objects, it still misclassified some small and rare categories (\eg rail). Second, we observe that the model can be easily confused by fine-grained categories that with similar semantic meaning. For example, the model sometimes wrongly classifies points of windowsill to the window class. 
In future work, it will be interesting to design models that can accurately distinguish between fine-grained categories in the open-vocabulary setting. 

\section*{Acknowledgements}
We sincerely thank Amir Hertz, Weicheng Kuo, and Lijun Yu for the helpful discussions. 

% \clearpage  % TODO REVIEW/FINAL: This \clearpage needs to be removed from both review and camera-ready versions.

\bibliographystyle{splncs04}
\bibliography{main}
\end{document}